\documentclass[runningheads]{llncs}

\usepackage[]{eccv}

\usepackage{eccvabbrv}

\usepackage{graphicx}
\usepackage{amsmath}
\usepackage{amssymb}
\usepackage{multirow}
\usepackage{booktabs}
\usepackage{comment}
\usepackage{here}
\usepackage{subcaption}

\usepackage{mlmath}
\usepackage[dvipsnames]{xcolor}
\newcommand{\red}[1]{{\color{red}#1}}
\newcommand{\cyan}[1]{{\color{cyan}#1}}
\newcommand{\green}[1]{{\color{green}#1}}

\newcommand{\comk}[0]{COM Kitchens\xspace}

\newcommand{\ca}[1]{\multicolumn{1}{c}{#1}}
\newcommand{\cl}[1]{\multicolumn{1}{c|}{#1}}

\usepackage[accsupp]{axessibility}  

\usepackage[breaklinks,colorlinks,citecolor=eccvblue]{hyperref}

\usepackage{orcidlink}

\begin{document}

\title{COM Kitchens: An Unedited Overhead-view Video Dataset as a Vision-Language Benchmark} 

\titlerunning{COM Kitchens}

\author{Koki Maeda*\inst{1,2}\orcidlink{0009-0008-0529-3152} \and
Tosho Hirasawa*\inst{1,3}\orcidlink{0000-0003-4657-8214} \and
Atsushi Hashimoto\inst{1}\orcidlink{0000-0002-0799-4269} \and \\
Jun Harashima\inst{4} \and
Leszek Rybicki\inst{4} \and
Yusuke Fukasawa\inst{4} \and
Yoshitaka Ushiku\inst{1}\orcidlink{0000-0002-9014-1389}}

\authorrunning{K.~Maeda et al.}

\institute{OMRON SINIC X Corp., Tokyo, Japan (*equally contributed) \and
Tokyo Institute of Technology, Tokyo, Japan \and
Tokyo Metropolitan University, Tokyo, Japan, \and
Cookpad Inc., Yokohama, Japan}

\maketitle
\begin{center}
 \centering
 \captionsetup{type=figure}
 \includegraphics[width=\linewidth]{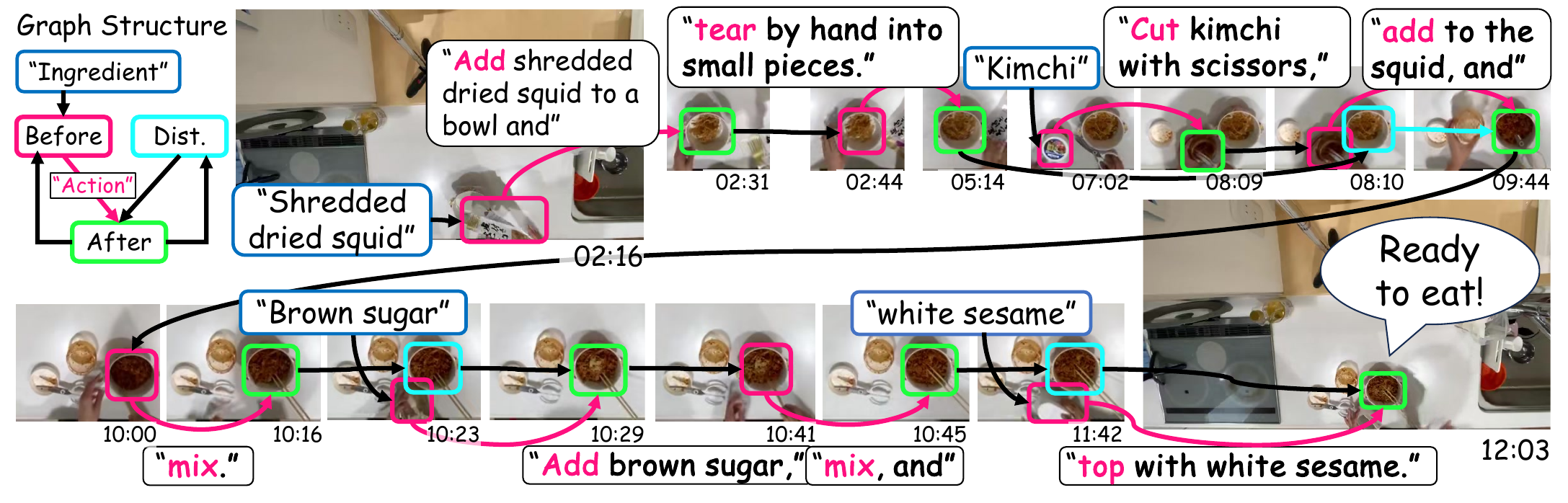}
 \captionof{figure}{
Sample from COM Kitchens: the dataset includes unedited overhead-view cooking videos, each manually annotated with a \textbf{visual action graph} that links instructional texts to visual elements through edges from {\it before} (\red{$\Box$}) to {\it after} (\green{$\Box$}) bounding boxes (BBs). {\it dist.} BBs (\cyan{$\Box$}) represent mixing. Details are provided in \cref{ss:def}.
  }
  \label{fig:visual_action_graph}
\end{center}%
\begin{abstract}   
Procedural video understanding is gaining attention in the vision and language community.
Deep learning-based video analysis requires extensive data.
Consequently, existing works often use web videos as training resources, making it challenging to query instructional contents from raw video observations.
To address this issue, we propose a new dataset, \textbf{COM Kitchens}.
The dataset consists of \textbf{unedited overhead-view videos captured by smartphones}, in which participants performed food preparation based on given recipes.
Fixed-viewpoint video datasets often lack environmental diversity due to high camera setup costs.
We used modern wide-angle smartphone lenses to cover cooking counters from sink to cooktop in an overhead view, capturing activity without in-person assistance.
With this setup, we collected a diverse dataset by distributing smartphones to participants.
With this dataset, we propose the novel video-to-text retrieval task \textbf{Online Recipe Retrieval} (OnRR) and new video captioning domain \textbf{Dense Video Captioning on unedited Overhead-View videos} (DVC-OV).
Our experiments verified the capabilities and limitations of current web-video-based SOTA methods in handling these tasks.
The dataset and code are available at 
\url{https://doi.org/10.32130/rdata.6.1} and 
\url{https://github.com/omron-sinicx/com_kitchens}, respectively.

\end{abstract}

\section{Introduction}
\label{sec:intro}
Creating higher-value products from raw materials is fundamental to material wealth in society.
Understanding these processes through video observation has gained increased attention from the computer vision community~\cite{kuehne2014cvpr,dean2017cvpr,Damen2018epic,ben2021ikea,nishimura2021ICCV,sener2022assembly101}.
Recent advancements in procedural video analysis rely on large-scale datasets collected from the Web~\cite{zhou2018youcook2,zhukov2019cvpr,miech19howto100m} or through ego-vision~\cite{gtea,Damen2018epic,nishimura2021ICCV,grauman2022ego4d}.
These datasets play a crucial role in developing robust techniques and pre-trained models~\cite{miech2020endtoend,Luo2020UniVL,Luo2021CLIP4Clip,Dvornik2023CVPR,lin2022egocentric,Ashutosh_2023_CVPR}.
Fixed-viewpoint observation was a major video format before the deep learning era~\cite{ace_dataset,50salads,kuehne2014cvpr}.
However, it has received little attention in the last decade because it is hard to find videos with such a format on the Web.
A cold-start problem exists: sufficient data is prerequisite to implement a raw-video-based retrieval system. Once available, the system encourages users to upload unedited videos for querying, reinforcing the dataset size.

To overcome this cold-start problem, we revisit unedited video datasets with fixed viewpoints (FV).
To efficiently enhance environmental diversity, we leveraged recent hardware advancements.
Modern smartphones are equipped with well-calibrated wide-angle cameras, allowing us to capture large work areas from an overhead view.
This way, we can observe processes with minimal occlusion regardless of kitchen layout, making it preferable for practical applications.
Additionally, people are familiar with smartphone UIs, allowing us to collect videos without in-person assistance.
To prove this concept of new-style dataset development, we scaled the FV procedural video datasets in this way, resulting in 145 videos, which are 40 hours in total footage, as the first collection.

Revisiting FV videos in this deep learning era offers challenges in understanding long procedural contexts rather than frame-wise image processing.
To tackle this, we provide a new dataset, \textit{COM Kitchens}, with a manually annotated visual action graph \cite{shirai2022visual}, linking visual events and text instructions with a workflow graph (Fig. \ref{fig:visual_action_graph}).
Using this structured annotation, we introduce a novel video2text retrieval task, \textit{online cross-modal recipe retrieval (OnRR)}, and a new domain for video captioning, \textit{dense video captioning on unedited overhead-view videos (DVC-OV)}.
OnRR is an online cross-modal task for retrieving corresponding recipes during cooking, designed to develop practical smartphone applications.
DVC-OV is an offline cross-modal task that generates instructional text from demonstrations, intended to analyze the domain gap between web- and overhead-view videos through a traditional format of the DVC task.


The contribution of this paper is four-fold.
\begin{enumerate}
\setlength{\itemsep}{0pt}
\item We introduce a novel approach to construct datasets of unedited fixed-view videos in diverse environments, leveraging modern smartphones.
\item We provide visual action graph annotations for the first time on unedited videos.
\item We propose a novel task of online recipe retrieval, including its target recipe pool and baselines.
\item We analyzed the SOTA dense video captioning method on unedited overhead-view videos to reveal future challenges.
\end{enumerate}

\section{Related work}

This paper proposes a new vision-language video dataset with one novel tasks and one novel domain for DVC.
We present comparisons to clarify the dataset novelty in \cref{ss:dataset}, and the task/setting novelties in \cref{ss:oxr}.

\subsection{Datasets for Procedural Video Understanding}\label{ss:dataset}

\begin{table}
  \caption{
    Comparison between instructional video datasets with fixed-viewpoint (FV) cameras. We categorized temporal segment type into \textit{action} (e.g., "put a bowl," "crack the egg," "beat the egg"), and \textit{step}, a higher-level action (e.g., "whisk eggs in a bowl").
  }
  \label{tab:datasets_fv}
  \centering
  \resizebox{\textwidth}{!}{
    \begin{tabular}{ll|lrrrrrcl}
      \toprule
        \ca{dataset} & \cl{year} & \ca{topic} & \ca{tasks} & \ca{\# env.} & \ca{\# videos} & \ca{total (h)} & \ca{avg. (m)} & \ca{seg. type} & \ca{seg. description} \\
      \midrule
        MMAC~\cite{mmac}                        & 2008 & Cooking & 1 & 1 & 32 & 8 & 15.0 & action & 130 actions cls.\\
        MPII~\cite{mpii}                        & 2012 & Cooking & 14 & 1 & 44 & 8 & 13.4 & action & 65 actions cls. \\
        ACE~\cite{ace_dataset}                  & 2012 & Cooking & 5 & 1 & 35 & 2 & 3.6 & action & 8 actions cls. \\
        50 salads~\cite{50salads}               & 2013 & Cooking & 2 & 1 & 50 & 5 & 5.4 & action & 51 actions cls. \\
        Breakfast~\cite{kuehne2014cvpr}         & 2014 & Cooking & - & 18 & 1,712 & 77 & 2.7 & action & 10 actions cls. \\
        IKEA ASM~\cite{ben2021ikea}             & 2021 & Furniture & 4 & 5 & 371 & 35 & 5.7 & action & noun+verb (n+v) \\
        Assembly101~\cite{sener2022assembly101} & 2022 & Assembly & 15 & 1 & 4,321 & 513 & 7.1 & act./step & 1,380 act. cls./n+v \\
        \textbf{COM Kitchens}                   & Ours & Cooking & 139 & 70 & 145 & 40 & 16.6 & act./step & instructional text \\
      \bottomrule
    \end{tabular}
  }
\end{table}

\cref{tab:datasets_fv} summarizes datasets with FV procedural videos.
They all target manufacturing tasks.
Among these datasets, COM Kitchens has a significantly \textbf{large diversity in tasks and environments} and is the only one with \textbf{linguistic annotations}.
We omitted the  EgoExo4D~\cite{grauman2023egoexo4d} dataset from the table because the paper does not provide organized statistics.

Breakfast~\cite{kuehne2014cvpr}, EgoExo4D~\cite{grauman2023egoexo4d}, and our dataset have environmental diversity, while others struggle with the cost of in-person technical support for their setup.
The Breakfast dataset collected data from 18 environments (overhead or side view), but its tasks are limited to two salad recipes with only 10 action classes.
EgoExo4D addressed the high cost by collaborating with 12 institutes.
Our dataset is competitive with EgoExo4D in terms of the number of environments and participants for the cooking scenario but offers more variety in tasks (i.e., recipes).
Another difference is camera views. EgoExo4D used multiple cameras from the front, side, and back view angles, aiming to capture activities beyond the kitchen counter to bridge the gap between egocentric and egocentric views.
In contrast, our setup captures detailed food manipulations at the counter with minimal occlusions from the overhead view.


\begin{table}
  \centering
  \caption{
    Comparison between instructional vision-language video datasets. Only our dataset uses a fixed viewpoint. \textit{Coarse instruction} minimally describes steps with verb(s) and noun(s) in a YouCookII style \cite{zhou2018youcook2} (e.g., "tier dried squid"), whereas \textit{fine instruction} comes from real instructional texts (e.g., "tear dried squid by hand into small pieces").
    'manual*' annotates only the start of intervals.
  }
  \label{tab:datasets_vl}
  \resizebox{\textwidth}{!}{
    \begin{tabular}{ll|clrrrrllr}
      \toprule
        \ca{dataset} & \cl{year} & \ca{type} & \ca{topic} & \ca{tasks} & \ca{\# videos} & \ca{total (h)} &    \ca{avg. (m)} & \ca{seg. description} & \ca{interval} & \ca{\# seg.} \\
      \midrule
        YouCookII~\cite{zhou2018youcook2}       & 2018 & Web & Cook. & 89 & 2,000 & 176 & 5.3 & coarse instruction & manual &4,325\\
        ProceL~\cite{elhamifar2019unsupervised} & 2019 & Web & Multi. & 12 & 720 & 47 & 3.9 & coarse instruction & manual & 498\\
        COIN~\cite{coin}                        & 2019 & Web & Multi. & 180 & 11,827 & 476 & 2.4 & coarse instruction & manual & 46,354\\
        CrossTask~\cite{zhukov2019cvpr}         & 2019 & Web & Multi. & 83 & 4,700 & 376 & 4.8 & coarse instruction & manual & 19,278\\
        MMAC-Captions~\cite{mmac_caption}       & 2021 & Ego & Cook. & 5 & 146 & 16 & 13.4 & coarse instruction & manual & 5,002\\
        Epic Kitchens~\cite{Damen2022RESCALING} & 2022 & Ego & Cook. & 70 & 700 & 100 & 8.6 & narration & utterance & 39,596\\
        Ego4D~\cite{grauman2022ego4d}           & 2022 & Ego & Open & - & - & 3,670 & - & narration & manual* & - \\
        BioVL2~\cite{taichi2022jnlp}            & 2022 & Ego & Bio. & 5 & 32 & 3 & 5.3 & fine instruction & manual& 408\\
        VRF~\cite{shirai2022visual}             & 2022 & Web & Cook. & 200 & 200 & 2 & 0.7 & fine instruction & manual& 3,705 \\
        FineBio~\cite{finebio} & 2024 & Ego & Bio. & 7 & 226 & 14.5 & 3.9 & fine instruction & manual & 3,541\\
        \textbf{COM Kitchens} & Ours & FV & Cook. & 139 & 145 & 40 & 16.6 & fine instruction & manual & 2,852 \\
      \bottomrule
    \end{tabular}
  }
\end{table}

\cref{tab:datasets_vl} compares our dataset to other procedural video datasets with linguistic annotations.
We omit video datasets without manual annotations, such as HowTo100M~\cite{miech19howto100m} and YT-Temporal-1B~\cite{Zellers_2022_CVPR}, as they are for pre-training, not for downstream tasks.
The pioneering work of YouCookII~\cite{zhou2018youcook2} provides linguistic annotations of coarse instructions (e.g., "whisk egg, flour"), and many other works follow this manner.
Epic Kitchens and Ego4D have narrations as their linguistic annotation, but they tend to describe the details of each action (e.g., action name and target objects), which is still similar to coarse instruction.
While captioning for procedural image sequences~\cite{chandu2019acl,nishimura2019inlg,nishimura2020ieee} uses commercial recipe sites as the dataset resource for generating fine instructions (e.g., "Beat the egg whites with a mixer, starting on low speed"), there is a gap with current video datasets' linguistic resource.

BioVL2~\cite{nishimura2021ICCV}, FineBio~\cite{finebio}, and VRF~\cite{shirai2022visual} are datasets with fine instructions, as this work.
BioVL2 is a rare dataset capturing biochemical experiments, but it is limited in size. FineBio is a five times larger than BioVL2 in total footage, but its task variation is still limited and the average footage is shorter than general cooking tasks.  
VRF collected one-minute videos focused on food state changes, excluding human actions. 
COM Kitchens has 70\% of the diversity against the web-based VRF dataset, consisting of unedited videos with 16.6 minutes on average.

Our dataset provides its annotation as \textbf{visual action graphs}.
This structured representation of manufacturing instruction was first proposed in 1980 by Momouchi~\cite{momouchi1980coling} as an estimation target for a natural language processing task.
Later, \textit{Flow graph}~\cite{mori2014lrec,yamakata2020lrec} was proposed with a fine-grained graph as a comprehensive representation of understanding.
\textit{Merging tree} was the simplest structure of the manufacturing process~\cite{jermsurawong2015emnlp}, which selects only actions that merge materials as nodes~\cite{jermsurawong2015emnlp}.
\textit{Action graph} is an intermediate representation between flow graph and merging tree, represents both actions on single materials and merging actions~\cite{kiddon-etal-2015-mise}.

These structures have been independently extended to vision-language setups.
\textit{Visual action graphs} were defined as a prediction target in an unsupervised task~\cite{dean2017cvpr} using web videos.
Unfortunately, their visual action graph dataset is only for testing and is not publicly available.
A small dataset of visual merging trees has been provided in~\cite{nishimura2020ieee} for semi-supervised learning, where visual data consists of image sequences instead of videos.
VRF~\cite{shirai2022visual} provides visual flow graphs with videos, but the videos are only one minute long.
COM Kitchens is the only dataset that annotates graph structures for action segments.
Note that the graph represents dynamic changes across frames, including merging and splitting processes. This point is an essential difference from video scene graph datasets~\cite{vidor,Yang_2023_CVPR}.

\subsection{Video-Text Retrieval and Video Captioning}\label{ss:oxr}

Retrieval is one of the fundamental tasks for cross-modal problems.
For videos, video-text retrieval~\cite{Luo2020UniVL,Luo2021CLIP4Clip,Ma2022XCLIP,Chen_2020_CVPR}, a task to find the video whose entire contents fit a text query, is the principal retrieval task.
The literature often evaluates the video-to-text scenario in addition to the text-to-video condition; web videos usually have text metadata, making video-to-text evaluation less practical.
Instead, the OnRR task assumes retrieving web content by its text from raw video observation. Since our dataset is sourced from smartphones, developed techniques should be directly applicable to smartphone videos.

Video paragraph captioning is a video captioning task specially designed for procedural videos; it assumes that event segments are given~\cite{lei2020acl,shi2020acmmm,nishimura2021acmmm,wu2022icmr}. However, for unedited videos, it is not practical to assume such given event segments.
Hence, we focus on dense video captioning (DVC)~\cite{johnson2016densecap,zhou2019cvpr}, a joint task of event detection and event description generation, as a fundamental task of video comprehension.
The recent main challenge of DVC is to suppress redundant detection~\cite{fujita2020eccv}.
To overcome this problem, "detect-then-describe"~\cite{shi2019acl,deng2021cvpr} and "describe-then-detect"~\cite{wang2021iccv} approaches have been studied.
The current SOTA method of Vid2Seq~\cite{yang2023vid2seq} describes captions and detects events in one stage by outputting segments as a time token with its caption.
Its performance is supported by the largest video dataset of YT-Temporal-1B~\cite{Zellers_2022_CVPR}.

We test our dataset with Vid2Seq to investigate the domain gap between Web and FV videos. The most significant gap is in video length and repetitive actions.
\cref{tab:datasets_vl} shows that videos in COM Kitchens are about three times longer than usual web videos on average (and 20 times longer than VRF, which are the TikTok style). This difference is mainly due to repetitive actions, often eliminated in web videos.
Another gap is the location of event-related objects within each frame. In web videos (or ego vision), the object of interest tends to be in the center of the frame, whereas FV videos do not dynamically focus on objects.
These repetitions and lack of focus on important objects provide additional challenges for aligning linguistic instructions to video events.

\section{The COM Kitchens Dataset}

Data collection and annotation are pivotal in determining the dataset's utility.
This section presents our data collection approach (\cref{ss:collect}), a detailed definition of the \textbf{visual action graph} (\cref{ss:def}), annotation procedure (\cref{ss:proc}), and dataset statistics (\cref{ss:stats}).

\subsection{Collecting Videos with their Corresponding Procedural Text}\label{ss:collect}

We selected candidate recipes for filming the cooking process from the Cookpad Recipe Dataset (CRD)~\cite{harashima-2016-large}.
The CRD is a comprehensive Japanese home cuisine database, including over 1.7 million recipes with ingredient lists.
In the selection process, we prioritized recipes estimated to take under 30 minutes to prepare and to have a moderate complexity level.
We excluded recipes using off-counter facilities (e.g., microwaves) because such actions would be recognizable with sound-based event identification, which is not our focus.
The selected recipes are typically but not limited to European, Chinese, and Japanese cuisines (with a certain level of localization).

To collect videos, we hired 110 participants from August to October 2021.
Each participant recorded up to four recipes at home.
Instead of in-person guidance, we provided an instructional document and video that directed the way of recording (e.g., cooking alone, where to capture, camera mode, and turning off TVs) and how to exclude privacy-related information (e.g., do not have a conversation with family, set the camera view not to capture the actor's face, and remove any private documents from camera view).
Participants signed a consent form acknowledging that the data would be used publicly for academic purpose. Participants received appropriate financial compensation.

We utilized the iPhone 11 Pro fixed to a tripod and recorded videos using the rear camera.
We instructed participants to set the camera to 30 fps, Full HD resolution of 1920$\times$1080, and ultra-wide field of view mode (equivalent to a focal length of 13 mm).
Despite the ultra-wide angle setting, there was almost no distortion; therefore, we applied no extra intrinsic calibration.
Appendix B provides filmed video examples.

Out of the 410 videos recorded, we excluded any that ignored the instructions, curating a total of 210 videos.
See Appendix C for detailed statistics. The most common reason for rejection was wrong observation areas. The second was unintentionally observed faces of actors.
We made our best effort to avoid involving private information, considering that participants are not professionals and the environment is personal.
This emphasis on privacy made data collection more challenging.

Among the 210 available videos, we have annotated 145 videos, which amount to a total of 40 hours. We will publish the remaining 85 videos as an unsupervised resource.
Two videos were recorded at 60fps; however, we retained them in the dataset as they exhibited no quality issues.
In the experiment, we used videos scaled to 640$\times$480 resolution, but we will release the original resolution together publicly.

\begin{figure}[t!]
  \centering
  \captionsetup{type=figure}
  \includegraphics[width=\linewidth]{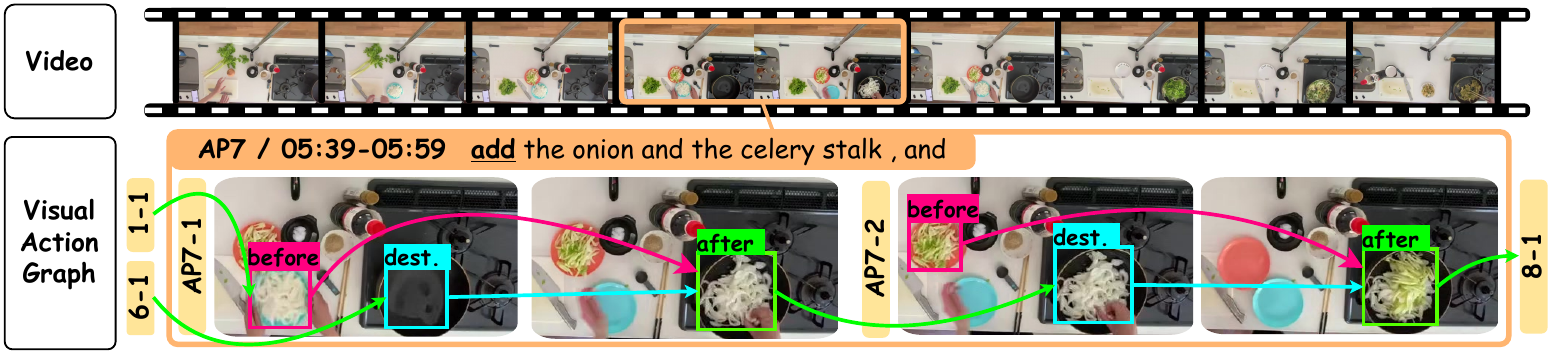}
  \captionof{figure}{
    A partial view of our visual action graph: AP7 consists of two sub-APs (7-1 and 7-2). All bounding boxes (BBs) mark foods (e.g., the destination BB in AP7-1 is {\it oil} heated in AP6-1). The duration is defined by the first and last BBs of the sub-APs.
  }
  \label{fig:teaser}
\end{figure}%

\subsection{Grounding Text to Video via Visual Action Graph}\label{ss:def}
A visual action graph delineates intentional actions termed \textbf{action-by-person (AP)} into a temporal graph format.
Here, AP (a.k.a., action-by-chef \cite{mori2014lrec}) refers to human actions on ingredients, like cutting or mixing, as opposed to natural processes, such as the browning of ingredients, which occur without human intervention.
We first present the formal definition of the visual action graph, followed by an illustrative example to contextualize its concept.

Define $\fV$ as the set of videos in our dataset, and each video $v \in \fV$ is a sequence of $|v|$ images, denoted $v = \{ v^{1}, v^{2}, \ldots,  v^{|v|}\}$.
Each video is paired with a corresponding sequence of procedural text, denoted as $t = \{ t^{1}, t^{2}, \ldots,  t^{|t|}\}$.
We define a visual action graph for a pair of video and text as $G(v, t) = (A, E)$, where $A$ symbolizes the set of APs and $E$ the set of edges, respectively.

Let $a_k$ be the $k$-th AP in the procedural text. 
We tie $a_k$ to visual content for encompassing the time segment in which the AP is performed and bounding boxes showing movements of objects, which can be denoted as
\begin{align*}
    a_k = (w_k, A_k, s_k, f_k).
\end{align*}
$w_k$ is a word sequence of the AP (e.g., $w_7=$``add the onion and the celery stalk, and" in \cref{fig:teaser}).
Due to the nature of the videos, $a_k$ may involve multiple object instances and multiple action instances.
$A_k$ represents them as a set of actions in the video (e.g., AP7-1 and AP7-2 for $a_7$).
We denote the $r$-th element of $a_k$ as $a_k^{\left(r\right)}=(b_{k}^{\left(r\right),\mathrm{bef}}, b_{k}^{\left(r\right),\mathrm{aft}}, b_{k}^{\left(r\right),\mathrm{dest}})$ (e.g., $a_7^{(1)}=$AP7-1), representing bounding boxes of the target ingredient \textbf{before}/\textbf{after} the action (\red{$\Box$} / \green{$\Box$}), and the \textbf{destination} (\cyan{$\Box$}) where the ingredient in $b_{k}^{\left(r\right),\mathrm{bef}}$ is mixed by the action. Note that $b_{k}^{\left(r\right),\mathrm{dest}}$ is not mandatory for $a_k^{(r)}$.
$s_k$ and $f_k$ denote the start and finish frame of the AP, defined by the first and last bounding box in $A_k$ (e.g., $s_7=05:39$ and $f_7=05:59$).

Nodes in the visual action graph are bounding boxes in APs defined above. 
Edges $E$ track ingredients throughout the cooking activity.
Here, $E$ consists of intra-action edges and inter-action edges.
Intra-action edges connect $b_k^{(r),\mathrm{bef}}$ to $b_k^{(r),\mathrm{aft}}$ (\red{$\Box$$\rightarrow$}\green{$\Box$}) and $b_k^{(r),\mathrm{dest}}$ to $b_k^{(r),\mathrm{aft}}$ (\cyan{$\Box$$\rightarrow$}\green{$\Box$}).
They are always within the same action, and we can automatically identify these edges based on the shared action index, $(k,r)$, where these indices are labeled instead of object names for this annotation.
In contrast, we manually identified inter-action edges, which is from $b_k^{(r),\mathrm{aft}}$ to $b_{k'}^{(r'),\mathrm{bef/dest}}$ (\green{$\Box$$\rightarrow$}\red{$\Box$} or \cyan{$\Box$}).

To automate the intra-action edge annotation, we labeled each bounding box with an action index instead of ingredient names.
Thus, the graph has no ingredient information in this form. To fix this problem, we manually gave ingredient names as leaf nodes and connected them to the AP that first processes the ingredient (ingredient-action edges), following the VRF dataset~\cite{shirai2022visual}.
This method naturally represents the composition of ingredients at each node by tracing back the edges to the leaf nodes.

\begin{figure}
  \centering
  \begin{subfigure}{0.49\textwidth}
    \includegraphics[width=\textwidth]{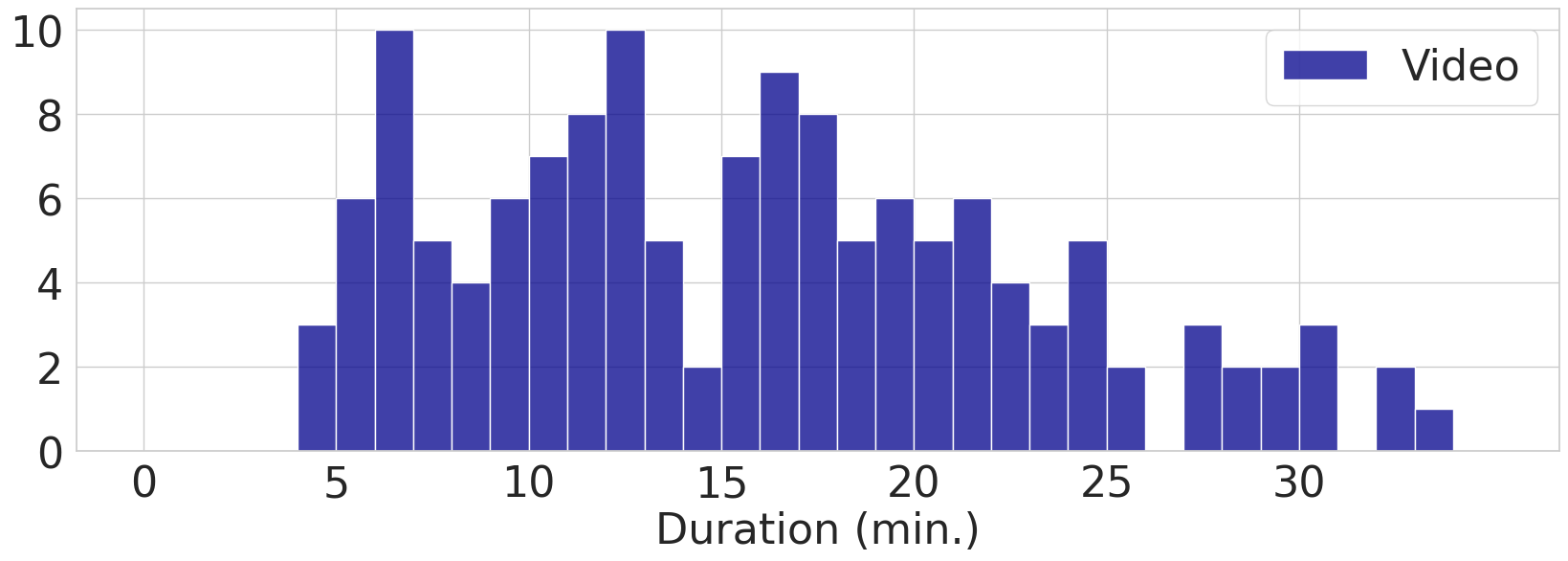}
  \end{subfigure}
  \hfill
  \begin{subfigure}{0.49\textwidth}
    \includegraphics[width=\textwidth]{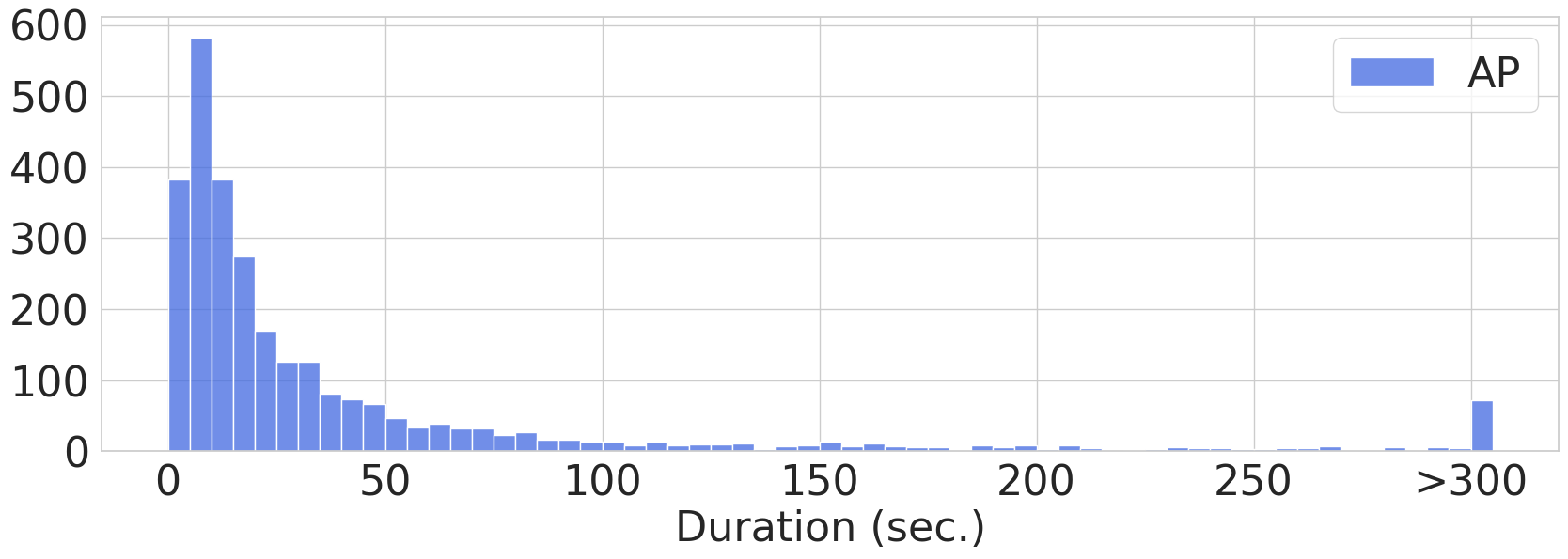}
  \end{subfigure}
  \caption{
    Distributions of duration; the averages are 16.6 min and 46.7 sec. for videos and APs, respectively.
  }
  \label{fig:duration}
\end{figure}


\subsection{Annotation Procedure} \label{ss:proc}
The visual action graph defined above was annotated for each video from February to September 2023.
A domain specialist annotated the graph to ensure consistency, with one of the authors reviewing it.
The annotator completed the task in about 430 hours.

Initially, the annotator reviewed the videos and revised the procedural instruction texts based on the performance in each video.
Despite instructions to follow the recipes faithfully, the participants often deviated due to the complexity of the cooking task. Instead, we used the revised recipes to simulate participants following them.
Simultaneously, the annotator identified and tagged the APs within the instructions.
All instructions were initially written in Japanese and translated into English by experts.
Each AP corresponded directly between the Japanese and English versions.

Subsequently, the annotator delineated the start and finish times with bounding boxes ($b_k^{(r),\mathrm{bef}}$, $b_k^{(r),\mathrm{aft}}$, $b_k^{(r),\mathrm{dest}}$) for each action. 
We used the Computer Vision Annotation Tool (CVAT)\footnote{\url{https://github.com/opencv/cvat}} for this annotation.
The video was examined every five frames to identify clear images of the target ingredients. 

Finally, the annotator has assigned inter-action edges to the bounding boxes. Following the automatic generation of edges in intra-action edges, they construct visual action graphs.
Ingredients and ingredient-action edges have also been annotated together with inter-action edges.
A total of 6,826 bounding boxes and 8,061 relationships were annotated to create the visual action graphs.

\begin{figure}[t!]
  \centering
  \begin{subfigure}{0.49\textwidth}
    \includegraphics[width=\textwidth]{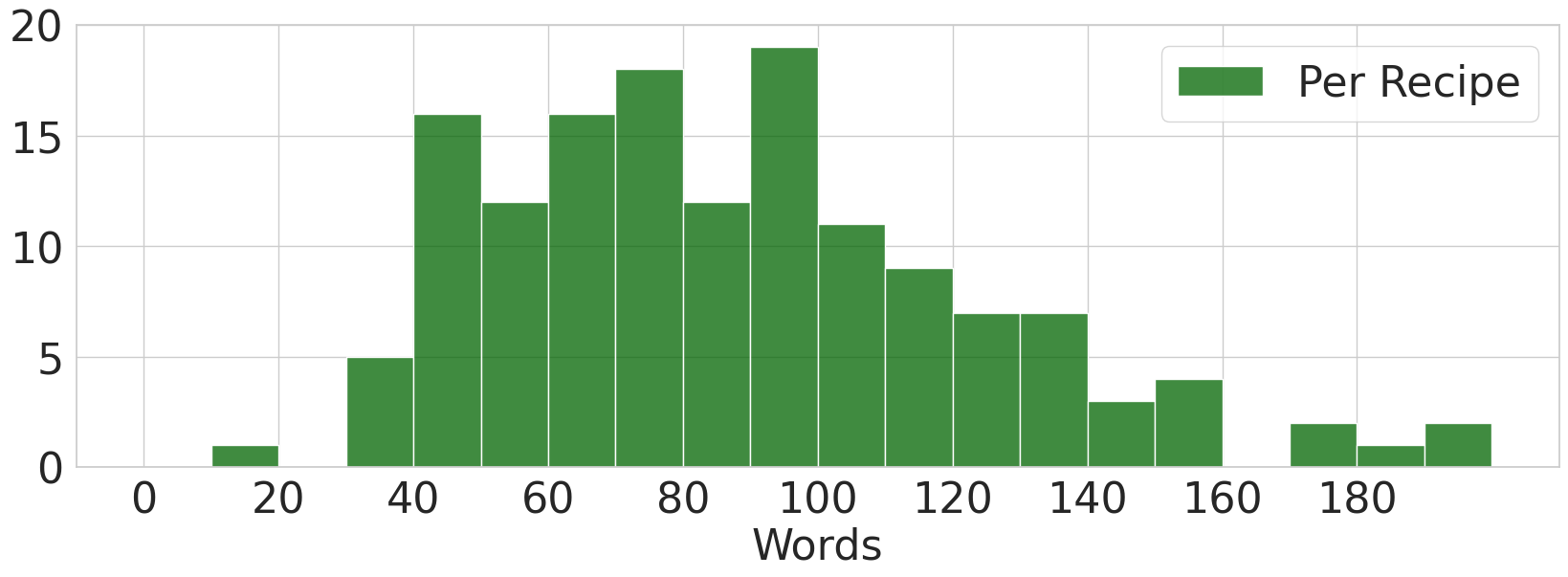}
  \end{subfigure}
  \hfill
  \begin{subfigure}{0.49\textwidth}
    \includegraphics[width=\textwidth]{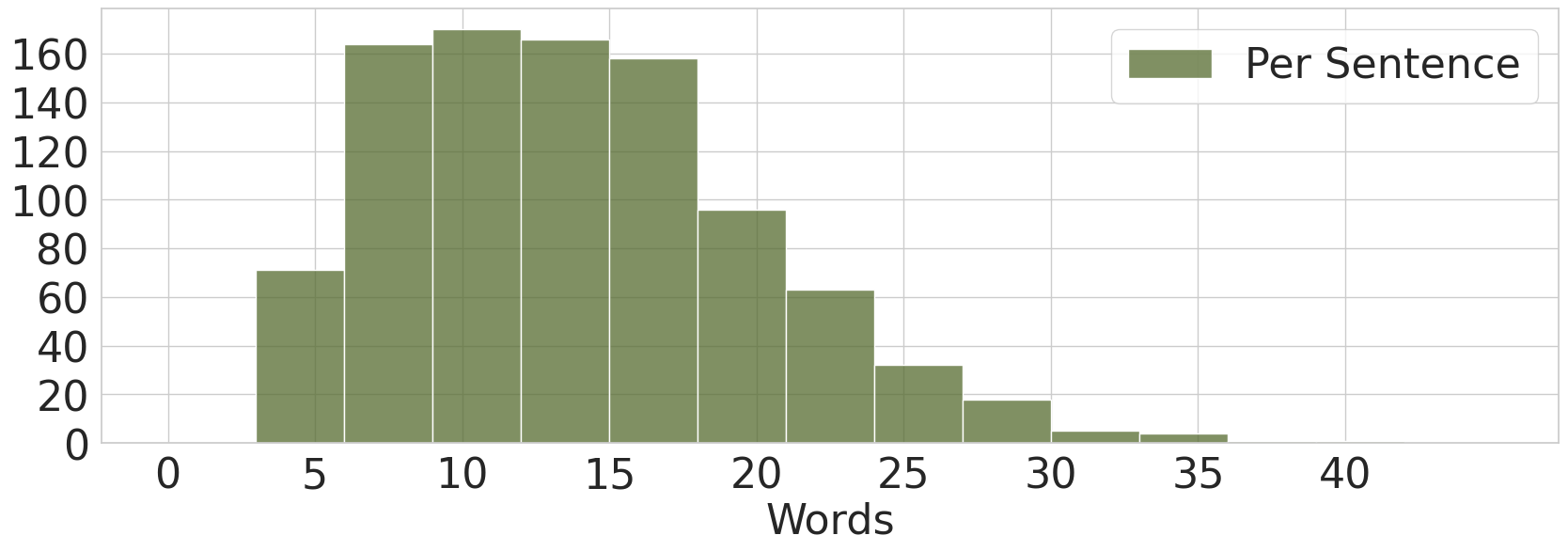}
  \end{subfigure}
  \caption{
    Distribution of the length of word sequences; the averages are 87.2 and 13.3 words for recipes and sentences.
  }
  \label{fig:word}
\end{figure}

\begin{figure}
  \centering
    \begin{minipage}[t!]{.48\linewidth}
        \centering
        \includegraphics[width=.90\linewidth]{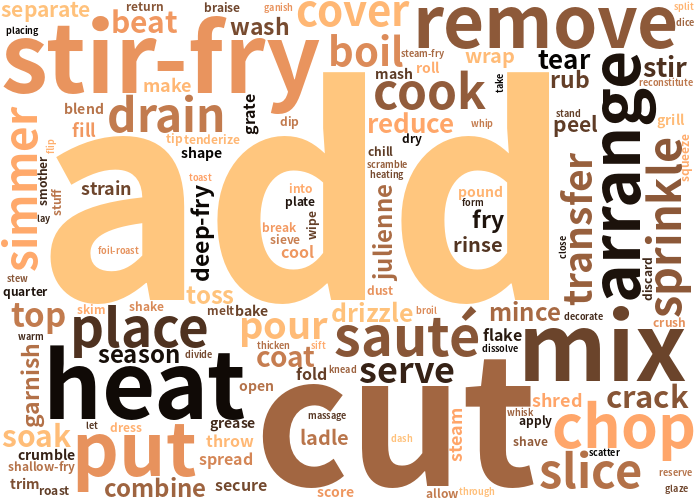}
        \includegraphics[width=.90\linewidth]{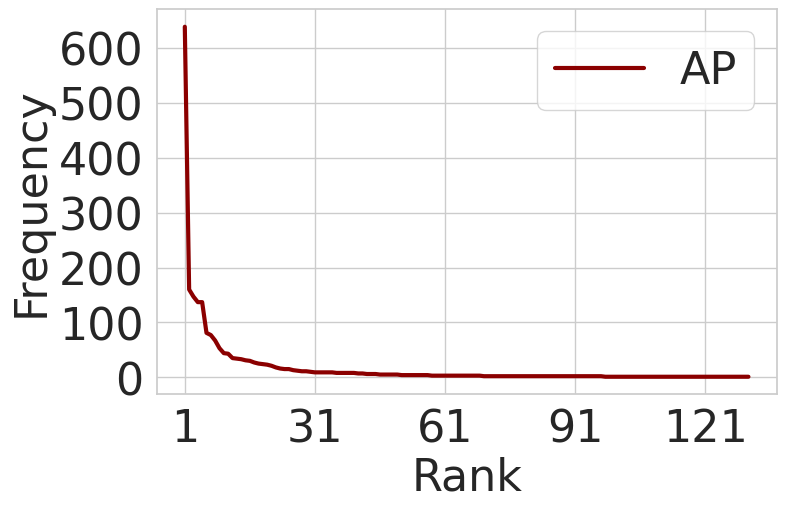}
        \subcaption*{AP}
    \end{minipage}\hfill%
    \begin{minipage}[t!]{.48\linewidth}
        \centering
        \includegraphics[width=.90\linewidth]{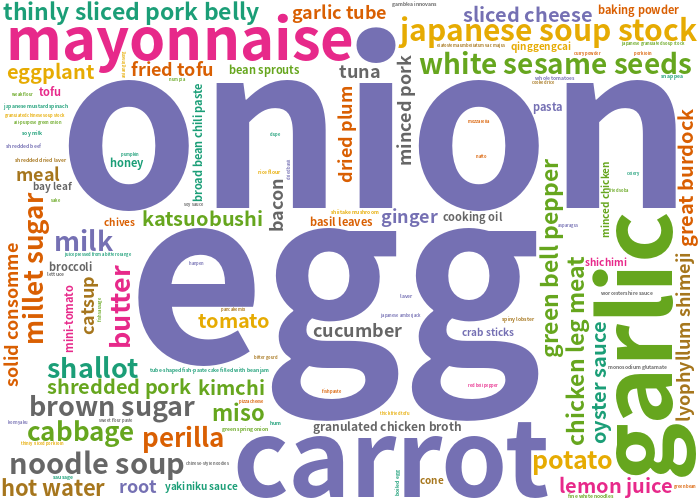}
        \includegraphics[width=.90\linewidth]{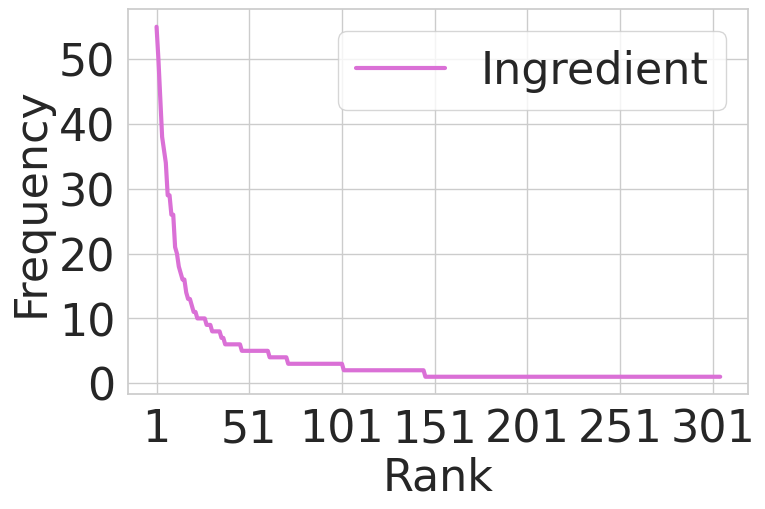}
        \subcaption*{Ingredients}
    \end{minipage}
  \caption{
    Word cloud and frequency plots. The word clouds visualize our target of fine-grained procedure comprehension well.
    There are 131 types of APs and 1,187 types of ingredients, both in a clear long-tail trend.
  }
  \label{fig:word-cloud}
\end{figure}

\subsection{Statistical Analysis}\label{ss:stats}

\paragraph{Recording Statistics.}
We aim to collect a large-scale dataset of fixed-viewpoint videos featuring structured, fine-grained annotations in diverse environments.
The COM Kitchens dataset includes 145 videos from 70 kitchens, totaling 40 hours.
Kitchens and actors are in one-to-one correspondence; providing 70 unique actors.
The average video duration is 16.6 minutes, as illustrated in \cref{fig:duration} (left).
80.0\% of the video frames are labeled with APs, with a 15.0\% overlap, which is comparable to other datasets\footnote{Referred to \cref{tab:datasets_fv} of the Assembly101~\cite{sener2022assembly101}.}.
On average, each AP lasted 46.7 seconds (\cref{fig:duration} right).
\cref{tab:datasets_fv} and \cref{tab:datasets_vl} show comparisons with similar recorded datasets.
Our dataset stands out for its diverse environments among fixed-viewpoint video datasets.
The average video length is longest among the datasets with linguistic annotations.

\paragraph{Recipe Data.}
Turning to the linguistic side, we assigned fine instructions comprising 949 sentences, averaging 6.5 sentences per video.
The average word lengths are 87.2 and 13.3 for recipes and sentences (\cref{fig:word}).
The instructions included 131 types and 2,286 distinct APs, averaging 15.8 per recipe and 2.4 per sentence.
An average of 1.24 repetitions was found in 14\% of the APs.
The top of \cref{fig:word-cloud} shows a wordcloud of AP, featuring terms beyond coarse instructions of other datasets in \cref{tab:datasets_vl}, including fine-grained cooking-specific words like ``Saute.''
The annotated instructions contained 305 types and 1,187 ingredients, averaging 8.2 per recipe.
The bottom graphs illustrate a distinct long-tail trend in AP and ingredients.

\begin{figure}[t!]
  \centering
  \includegraphics[width=0.9\linewidth]{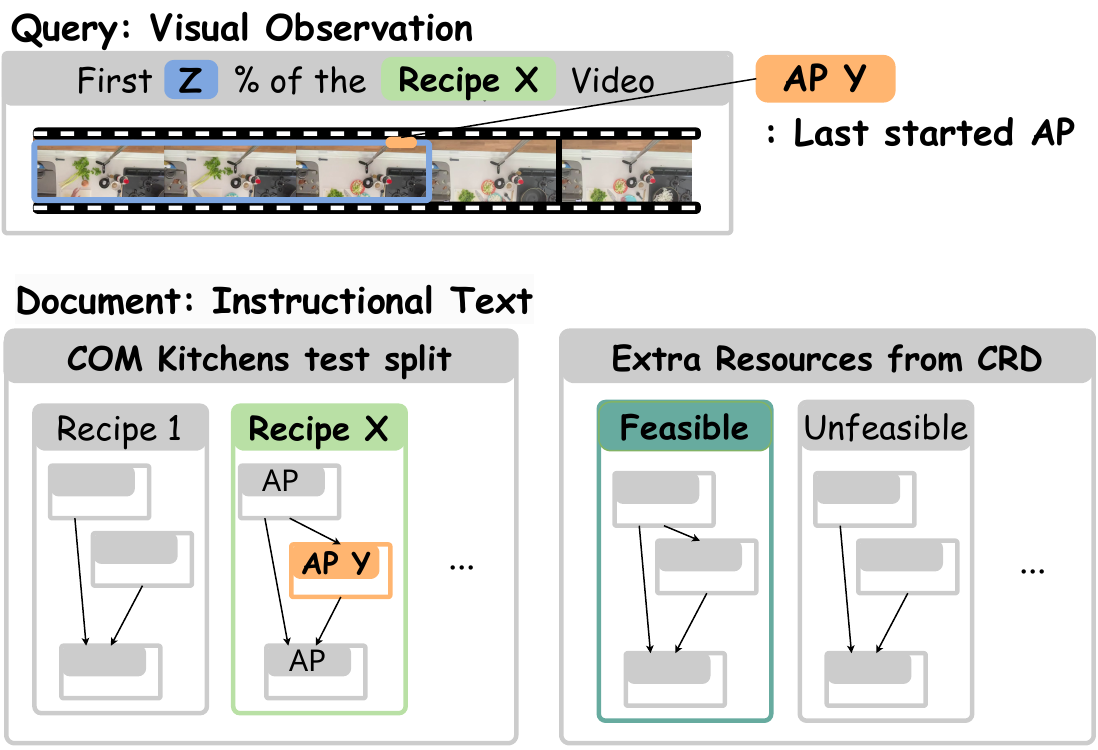}
  \caption{
    Data related to OnRR: The query of OnRR sub-tasks is the first $Z$\% of a video. For feasible recipe retrieval, we added an extra recipe resource to enhance the dataset of retrieval targets alongside our test set.
  }
  \label{fig:task-onrr}
\end{figure}

\section{Experiments}

\subsection{Online Recipe Retrieval}
Online Recipe Retrieval (OnRR) involves analyzing cooking videos up to the midpoint to determine the recipe type and the cooking stage achieved. 
\cref{fig:task-onrr} shows data related to this task; retrieving recipe texts that align with the video content up to a time point $Z$, using the video segment up to the last started AP $Y$ as the search query. This task is more challenging than traditional video-to-text retrieval as it requires both classification of video type and comprehension of the cooking process. 
We can decompose OnRR into two subtasks: {\it feasible recipe retrieval} and {\it recipe stage identification}. 
We assess these subtasks separately because solving them together is challenging for current SOTA methods.

Feasible recipe retrieval evaluates OnRR from a task categorization perspective in an online setting.
We consider a recipe feasible for a task in the middle of the workflow when an actor can shift to the recipe without disposing of foods. This setup assumes an application of online recipe recommendation.
Task success is measured by whether the retrieved recipe is in the feasible set.

Recipe stage identification is evaluated with a given recipe. The task is to spot the exact cooking stage in the recipe from the video. It involves aligning the video content with the corresponding part of the recipe text, up to the verb $w_y$ that matches AP $Y$, within a given recipe ID $X$. This subtask focuses on accurately matching the video content with the specific steps in the recipe.

\paragraph{Settings.}

In our experiment, $Z$ is set at 25\%, 50\%, 75\%, and 100\% from the initial point by frame. We labeled them early-, middle-, late-, and full-stage\footnote{The full-stage task is equal to conventional video-to-text retrieval.}, respectively.
The video set was split into 90/26/29 videos for the train/val/test sets, and the test set was further divided into $29\times 4=116$ stages.
We compared the performance of state-of-the-art models (UniVL~\cite{Luo2020UniVL}, CLIP4Clip~\cite{Luo2021CLIP4Clip}, and X-CLIP~\cite{Ma2022XCLIP}) against random selection.
We set the maximum token length, maximum frame length, batch size, and number of training epochs to 48, 48, 32, 5 for UniVL and 32, 12, 32, 3 for CLIP4Clip and X-CLIP.
These methods often cannot handle all the words in a recipe at once, as it often exceeds 100 words, as shown in \cref{fig:word}.
Hence, we shortened the recipe texts by POS tagging with spaCy, extracting only nouns and verbs to cover all steps with the 100 words.

\paragraph{Extra Recipe Resource.}
The number of retrieval targets should be sufficiently large; however, the COM Kitchens test set size is limited.
Thus, we have increased retrieval targets with an extra recipe resource from CRD.
First, we identified the ingredients appearing in the early stage ($Z=25$\%) by referring to the action graph for each test set video.
Based on the early-stage ingredients, we then extracted up to 100 candidate recipes from CRD for each test recipe. This operation collected 1,828 recipes as this extra resource.
Among them, we manually discovered 991/243/19/5 feasible recipes for each stage $Z$, tied with their corresponding test set recipe. 
Nonfeasible recipes remained as hard negatives, often overlapping in ingredients with feasible recipes.
Note that extracted recipes only include recipe texts and ingredients, not videos. We machine-translated them from Japanese to English \footnote{We used the DeepL API to translate from Japanese to English (US).}.

\paragraph{Metrics.}
Following the conventional video-text retrieval task, we employed Recall at rank K (R@K, higher is better) and median rank (MdR, lower is better) to evaluate the retrieval performances at the two subtasks.

\begin{table}
  \caption{
    Online recipe retrieval (OnRR) performances of baseline models. R@K and MdR represent recall at rank K $(\uparrow)$ and median rank $(\downarrow)$, respectively. This table provides only the early- and middel-stage settings (using the first 25\% and 50\% of the video as input); results in other stages and settings are detailed in Appendix D.
  }
  \label{tab:result_onrr}
  \centering
  \begin{tabular}{c|r|rrrr|rrrr}
    \toprule
      \multirow{2}{*}{Task} & \cl{\multirow{2}{*}{Method}}  & \multicolumn{4}{c|}{\textbf{Early (25\%)}} & \multicolumn{4}{c}{\textbf{Middle (50\%)}} \\
                            &                               & R@1 & R@5 & R@10 &  MdR & R@1 & R@5 & R@10 &  MdR\\
    \midrule
      \multirow{4}{*}{\begin{tabular}{c}
        Feasible \\ Recipe\\ Retrieval
      \end{tabular}}
        & Random                            & 1.8 & 8.6 & 15.8 & -    & 0.4 & 1.8 & 3.1 & - \\
        & UniVL~\cite{Luo2020UniVL}         & 3.4 & 5.7 & 9.2 & 227.0 & 3.4 & 5.7 & 9.2 & 231.0 \\
        & CLIP4Clip~\cite{Luo2021CLIP4Clip} & 0.0 & 0.0 & 10.3 & 79.0 & 0.0 & 0.0 & 6.8 & 85.0 \\
        & X-CLIP~\cite{Ma2022XCLIP}         & 0.0 & 6.8 & 10.3 & 89.0 & 0.0 & 3.4 & 3.4 & 320.0 \\
    \midrule
      \multirow{4}{*}{\begin{tabular}{c}
        Recipe \\ Stage \\ Identification
      \end{tabular}} 
        & Random                            & 6.3 & 31.6 & 63.3 & 8.0  & 6.3 & 31.6 & 63.3 & 8.0 \\
        & UniVL~\cite{Luo2020UniVL}         & 17.2 & 48.2 & 68.9 & 5.0 & 9.2 & 63.3 & 89.2 & 3.0 \\
        & CLIP4Clip~\cite{Luo2021CLIP4Clip} & 6.8 & 48.2 & 68.9 & 5.0  & 10.3 & 55.1 & 86.2 & 4.0\\
        & X-CLIP~\cite{Ma2022XCLIP}         & 10.3 & 51.7 & 68.9 & 4.0 & 17.2 & 37.9 & 93.1 & 6.0 \\
    \bottomrule
    \end{tabular}
\end{table}

\paragraph{Results.}

\cref{tab:result_onrr} showcases the result of the OnRR benchmark in the early- and middle-stage settings. Results for the other two stages and no-fine-tune settings are provided in Appendix D.
None of the baseline models outperformed random selection in feasible recipe retrieval. 
Conversely, all models outperformed random selection in recipe stage identification and improved with fine-tuning.
These results suggest that simple contrastive learning cannot solve these two tasks simultaneously.
In other words, the OnRR task serves as a benchmark for video-text retrieval based on a procedural comprehension.

\subsection{Dense Video Captioning on Unedited Fixed-viewpoint Videos}

Dense video captioning is another task for procedural videos, where a system generates multiple detailed captions for different segments within a video.
This approach involves detecting distinct events in the video timeline and then generating descriptive and accurate captions for each identified event.
The objective is to provide a more comprehensive and segmented understanding of the video content, which is beneficial for offline applications, such as accessibility, content analysis, and enhanced video search capabilities.

For the \comk dataset, we use APs as the segments of a video; the content is the recipe text of each AP and the entire duration covering all repetitions under each AP.
The main challenge of this task is the domain gap between our unedited overhead-view videos and traditional DVC targets such as web videos.

\paragraph{Settings.}
In this experiment, we used the same train/valid/test split with the OnRR task.
We selected two DVC systems as our baselines: PDVC~\cite{wang2021iccv} and Vid2Seq~\cite{yang2023vid2seq}.
We first tested their zero-shot performance on the \comk dataset.
Then, we fine-tuned the Vid2Seq model, the SOTA model on the YouCookII dataset, on our dataset and evaluated its performance.

Besides, we examined two supervision approaches to leverage the action graph in the DVC task: (i) action graph as relation labels (RL) and (ii) action graph as attention supervision (AS).
When we took the action graph as relation labels, we employed the decoder of TablERT-CNN~\cite{ma-etal-2022-joint} as our module to predict the relation (a.k.a. the type of edges) between APs and trained the model in a multi-task learning manner.
For attention supervision, following Garg \etal~\cite{garg-etal-2019-jointly}, we minimize the Kullback-Leibler divergence between the self-attention matrix at the last encoder layer and the alignment matrix indicating the existence of edges.

\paragraph{Metrics.}

We employed SODA\_c~\cite{fujita2020eccv}, CIDEr~\cite{vedantam2015cvpr}, and METEOR~\cite{banerjee2005acl} scores to evaluate model performance, as these are the commonly used for DVC tasks.

\paragraph{Results.}

\begin{table}[t!]
  \caption{
    Comparison between DVC performances of baseline models.
    The rows with `FT' of `\checkmark' show the results of models fine-tuned on \comk.
    ``AG'' shows the choice of methods to leverage action graph information during the fine-tuning.
  }
  \label{tab:dvc_results}
  \centering
  \begin{tabular}{@{}lccccc@{}}
    \toprule
        \multicolumn{1}{c}{Model} & FT & AG & SODA\_c$(\uparrow)$ & CIDEr$(\uparrow)$ & METEOR$(\uparrow)$ \\
    \midrule
        PDVC~\cite{wang2021iccv}       & - & - & 0.022 & 0.000 & 0.000 \\
        Vid2Seq~\cite{yang2023vid2seq} & - & - & 0.017 & 0.066 & 0.010 \\
    \midrule
        Vid2Seq & \checkmark & - & 0.369 & 2.832 & 0.642 \\ 
        Vid2Seq & \checkmark & RL  & 0.211 & 1.381 & 0.285 \\ 
        Vid2Seq & \checkmark & AS  & 0.266 & 2.513 & 0.423 \\ 
        Vid2Seq & \checkmark & RL$+$AS  & 0.581 & 6.195 & 1.142 \\ 
    \bottomrule
  \end{tabular}
\end{table}

\cref{tab:dvc_results} shows the result of the DVC benchmark.
The zero-shot performances of both the PDVC and Vid2Seq models are extremely low compared to those on the YouCookII dataset ($4.9$ and $7.9$ SODA\_c scores, respectively).
This deterioration indicates that the \comk dataset is challenging for the DVC task.
While speech information is proved to be a crucial modality to generate better captions~\cite{yang2023vid2seq}, videos in the \comk dataset have no speech information than web-based procedural video datasets.

\begin{figure}[t!]
  \centering
  \includegraphics[width=1.0\linewidth]{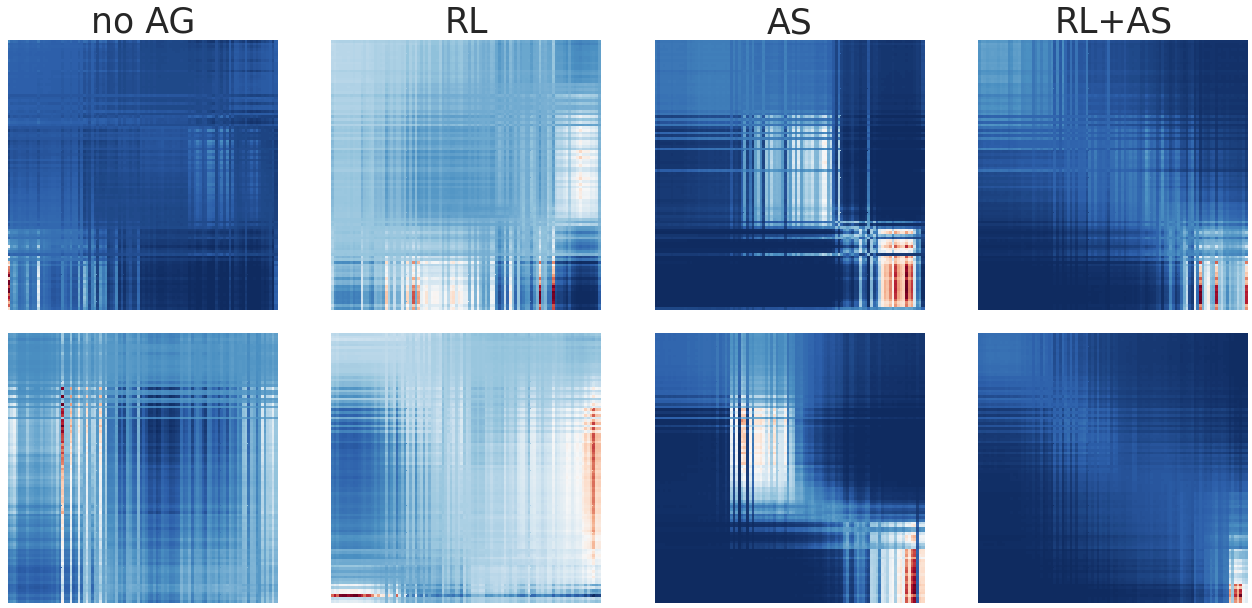}
  \caption{
    Visualization of the attention from the first head in the last encoder layer, based on randomly selected samples. Additional examples can be found in Appendix~E.
  }
  \label{fig:task-dvc-attn}
\end{figure}

We also found that the supervision obtained from the action graph benefits the model performance.
While RL or AS supervision individually does not bring improvement, their combination enhances model performance across all metrics.
This performance gain is attributed to improved attention.
As shown in the \cref{fig:task-dvc-attn}, RL or AS supervision aligns most frames to a small subset, but the RL+AS supervision aligns frames to their relative frames (defined by the action graph) and their surrounding frames.
This smoother alignment aids in generating recipe texts and determining segment boundaries.

\section{Discussion}
We provided baseline results for two specific tasks: OnRR and DVC-OV, extensions of conventional cross-modal retrieval and captioning tasks.
OnRR aims to create practical video comprehension for online applications.
In contrast, DVC-OV focuses on understanding overhead-view procedural videos as offline tasks.

A potential application involves constructing the visual action graph using video and recipe text inputs.
It aims to structurally understand cooking procedures by focusing on temporal dependencies and local human-object interactions.
Another possible task is episodic memory in the procedural domain, a linguistic query-driven reasoning task tailored originally for long-duration ego-centric videos~\cite{grauman2022ego4d}.
The COM Kitchens dataset is particularly suited for this task, as it comprises long videos (avg. 16.6 min., \cref{tab:datasets_fv}) and the visual action graph can generate various queries with spatio-temporal ground truth.

Our dataset, similar to many others, is limited by its size.
Furthermore, benchmark results indicate that conventional global alignment methods are ineffective for these tasks, prompting us to explore new pre-training and fine-tuning methods.
Fortunately, our data collection costs are significantly lower compared to datasets like EgoExo4D. 
We aim to further explore the potential of visual action graphs and to expand the dataset both with and without supervision.

\section{Conclusion}

We have introduced COM Kitchens, a dataset that facilitates vision-language understanding with overhead-view recordings, procedural recipe texts, and visual action graphs.
This dataset reflects real kitchen conditions, providing rich insights into the sequence of actions and states of ingredients.
Our experiments on the OnRR and DVC-OV benchmarks revealed the limitations of existing cross-modal retrieval models in handling long sequences and temporal dependencies. 
We plan to expand COM Kitchens for further challenges, believing that this dataset will contribute to the advancement of complex video content interpretation.

\section*{Acknowledgement}
This work would not have been possible without the meticulous and precise annotations by Yasuko Sonoda.
We are deeply grateful for her significant contributions to the dataset construction.
We also express our gratitude to the IDR office at the National Institute of Informatics for their continuous support in our dataset releases.
This work was supported by JSPS KAKENHI Grant Number 21H04910 and JST Moonshot R\&D Grant Number JPMJMS2236.



%
%
\bibliographystyle{splncs04}
\bibliography{main}

\newpage
\appendix
\section{Detailed Data Appendices}

To further aid in understanding, a few examples from \comk are provided with the supplementary material in the \path{examples} directory.
Samples include (i) unedited recorded videos, (ii) annotations for Japanese recipes, (iii) annotations for translated English recipes, and (iv) constructed visual action graphs. Besides, we also provide a video wall (\path{videowall.mp4}) to overview the unedited videos, which demonstrates the diversity of our dataset.

\section{Film set}\label{app:sec:film_set}

\begin{figure}[htbp]
    \centering    
        \begin{minipage}[b]{0.29\linewidth}
            \centering
            \includegraphics[width=\linewidth]{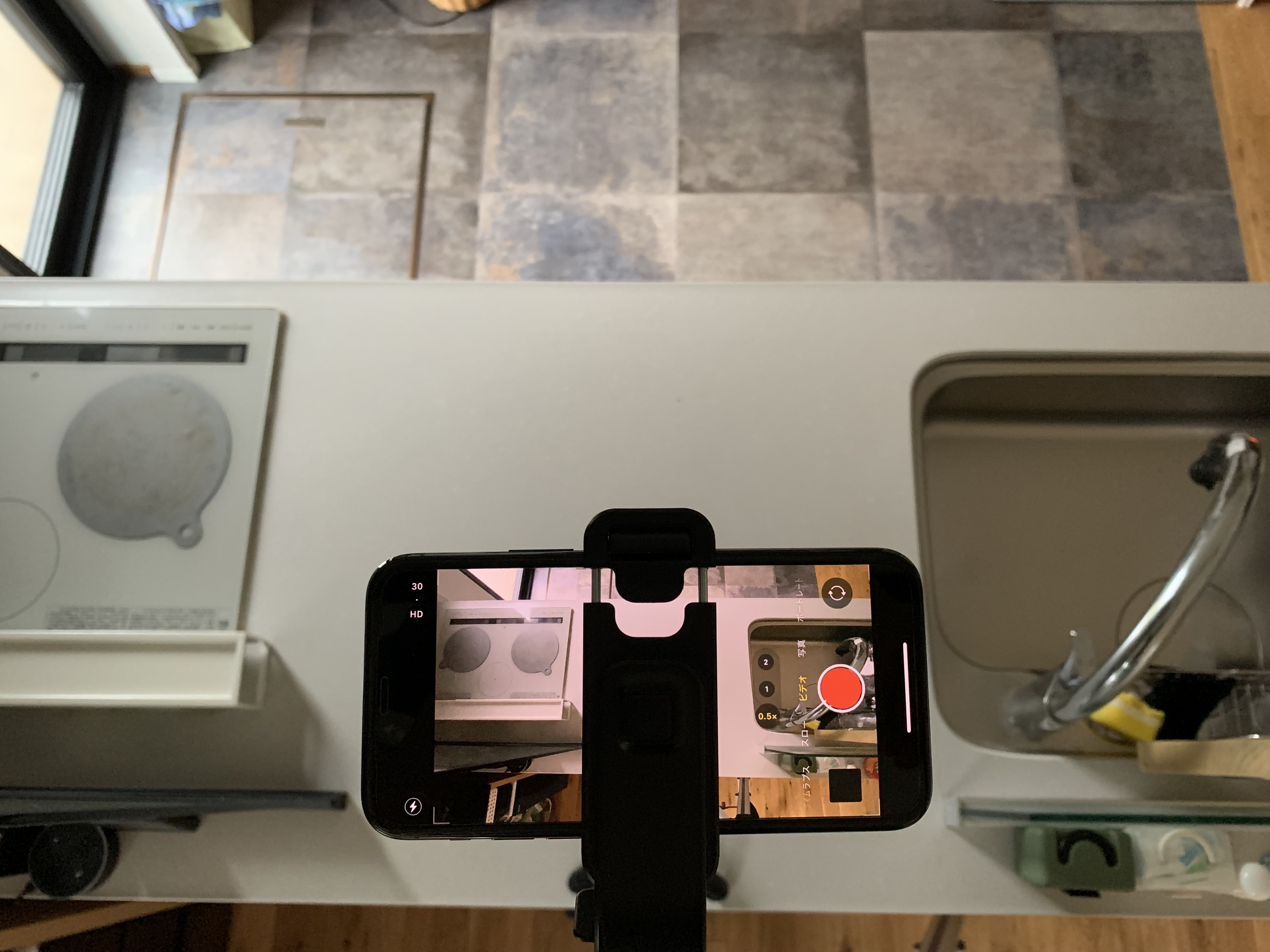}
            \subcaption{Top-down view.}
        \end{minipage}\hfill%
        \begin{minipage}[b]{0.29\linewidth}
            \centering
            \includegraphics[width=\linewidth]{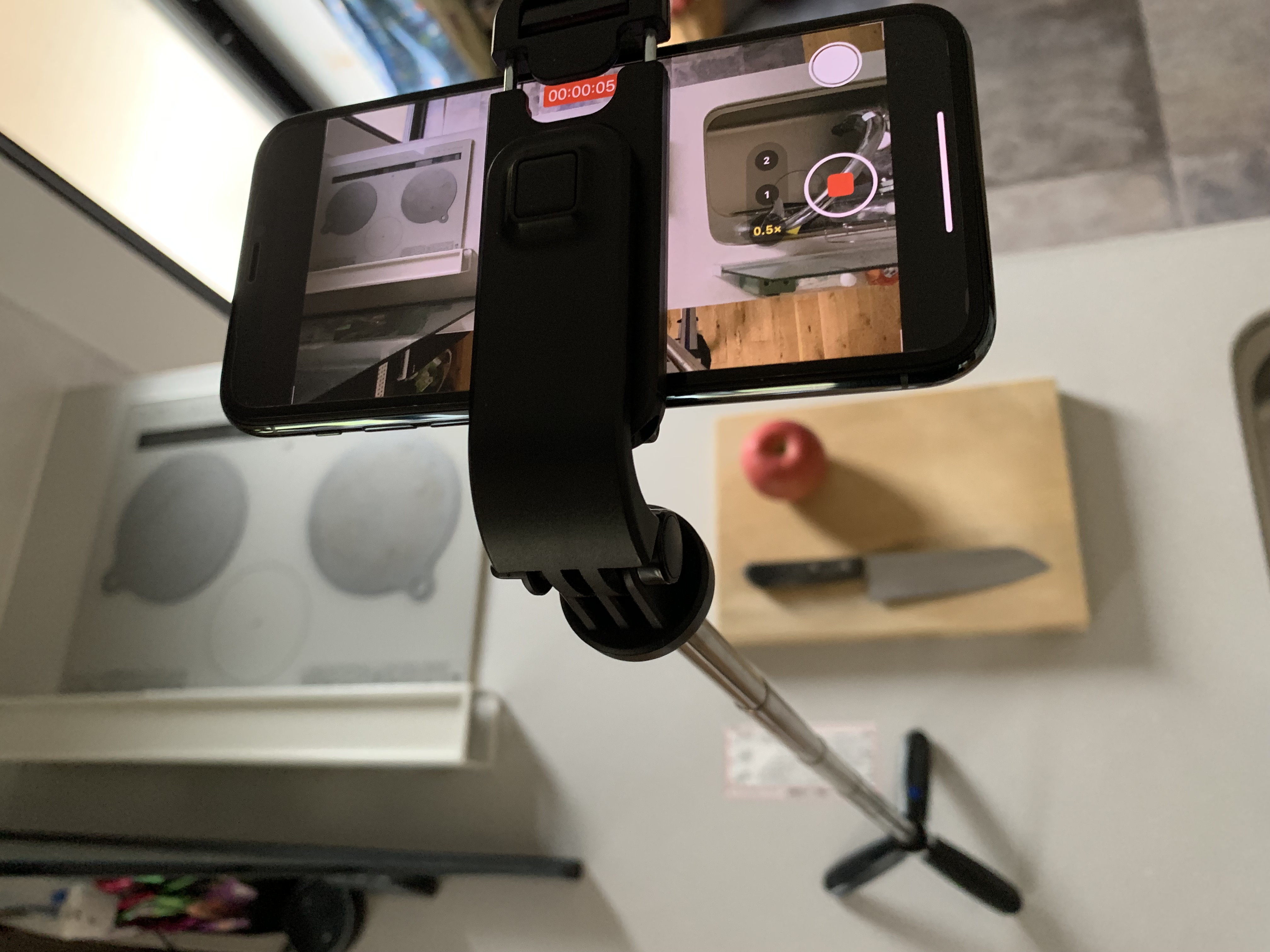}
            \subcaption{Angled top-down view.}
        \end{minipage}\hfill%
      \begin{minipage}[b]{0.385\linewidth}
        \centering
        \includegraphics[width=\linewidth]{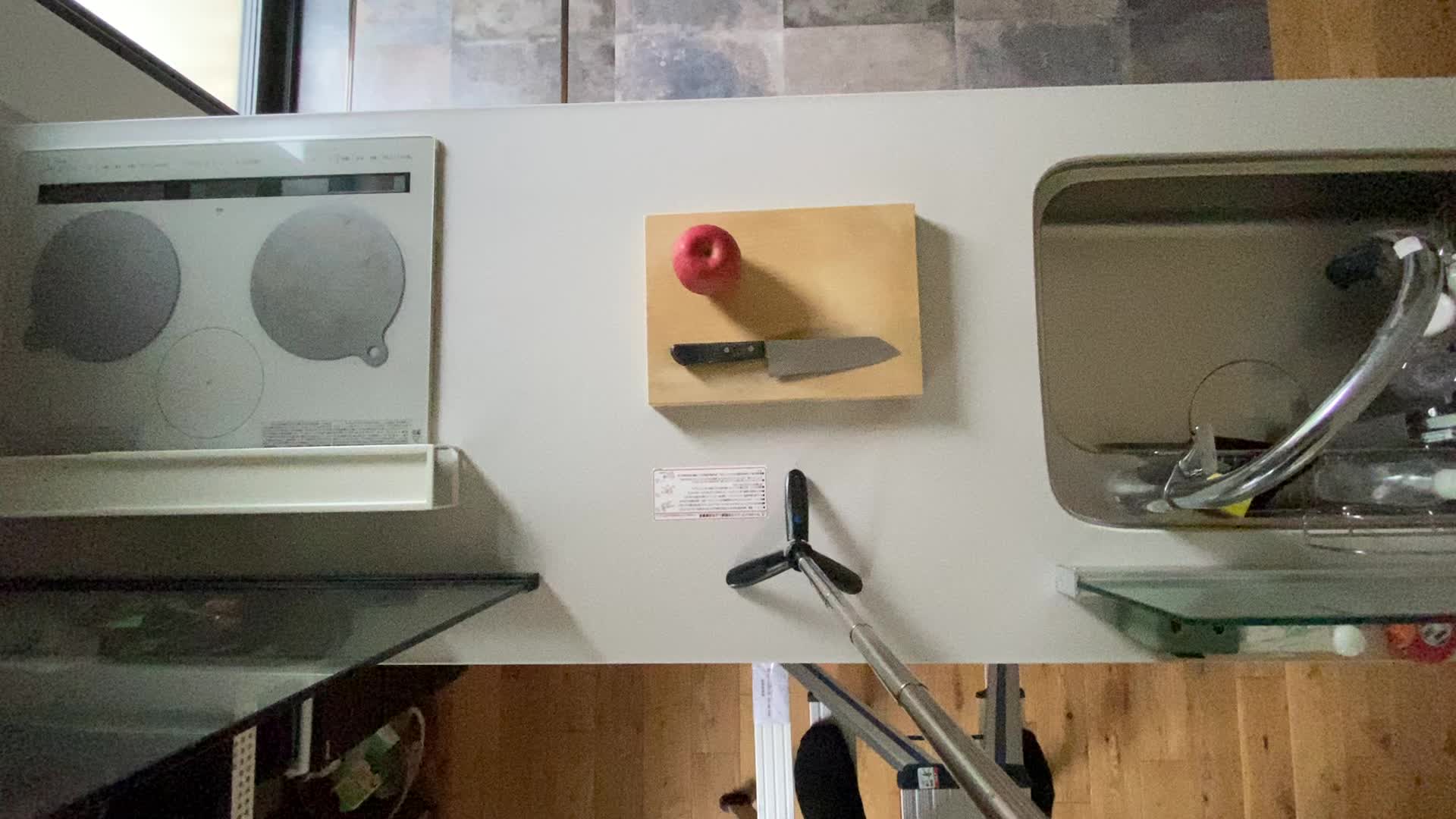}
        \subcaption{The footage of a capture.}
      \end{minipage}
    \caption{Example of the film set and recorded content.}
    \label{fig:env}
\end{figure}

We provide an example of the film set in \cref{fig:env}.
In the recording, we employed a tripod with 900 mm of height and instructed to place it with prior confirmation that the wide-angle mode of the rear camera could cover the whole kitchen top.


\section{Reason of rejection}\label{app:sec:rejection_statistics}
\begin{table}
\caption{Breakdown of reasons for refusal with statistics.}
\label{tab:reason_of_refusal}
\centering
\begin{tabular}{cl|rr}
    \toprule
        & \multicolumn{1}{c|}{Reason} & \multicolumn{1}{c}{\# of Refusals} & \multicolumn{1}{c}{\% among Refusals} \\
    \midrule
    1. & Inappropriate view (e.g., stove is not covered) & 97 & 50.5\% \\
    2. & Faces in the view & 46 & 24.0\% \\
    3. & Skipped steps using pre-processed food & 12 & 6.3\% \\
    4. & Overly complicated process & 11 & 5.7\% \\
    5. & Pause and resume in recording & 9 & 3.1\% \\
    6. & Recording by slow mode & 5 & 2.6\% \\
    7. & Personal documents in the view & 3 & 1.6\% \\
    8. & Any other reasons & 10 & 3.1\% \\
    \bottomrule
    \end{tabular}
\end{table}

We summarize the reason for refusal with its statistics in \cref{tab:reason_of_refusal}. We had to refuse roughly 50\% of the submitted videos (192/412), which is a relatively high rate. 
Our instructional videos and documents are almost for items 1 and 2, but the ignorance of those instructions caused 74.5\% of refusals.
This was caused primarily due to the lack of pre-filtering.
Since we selected to collect videos with the same smartphone model this time, we had to distribute our equipment to participants, which made pre-filtering difficult. 

We judged a procedure too complicated if the video duration was more than one hour or had more than 30 APs or 10 actions in an AP. They were caused by our failure in the recipe selection. 
In addition, we refused some videos if an actor repeated tasting and adjusting the taste too many times or repeated actions of wrapping small ingredients that were almost invisible in the video.

The other reasons were incomplete information in the consent form (3 videos), withdrawal of consent at the request (2 videos), and removal of the recipe from the Cookpad website (1 video). 


\section{Additional Results on the OnRR task}\label{app:sec:onrr}

\cref{tab:result_onrr_notrain} lists the result of baseline models in the OnRR benchmark in the early- and middle-stage setting. \cref{tab:result_onrr_stage} showcases the rest results, late-, and full-stage setting.
These results suggest that in the recipe stage retrieval task, fine-tuning with our \comk dataset improves the performance, regardless of model types and cooking stage settings.
On the other hand, the reduced and unstable performance in the feasible recipe retrieval task implies that the conventional contrastive learning strategy does not fit the objectives.



\begin{table}[htbp]
\caption{Online recipe retrieval (OnRR) performances of baseline models \textbf{without fine-tuning on \comk} in the early- and middle-stage settings (using the first 25\% and 50\% of the video as input).
R@K and MdR represent recall at rank K $(\uparrow)$ and median rank $(\downarrow)$, respectively. 
The results with fine-tuning are shown in \cref{tab:result_onrr}.}
\label{tab:result_onrr_notrain}    \centering
\scalebox{1.0}{
    \tabcolsep = 1.1mm
    \begin{tabular}{c|r|rrrr|rrrr}
    \toprule
    \multirow{2}{*}{Task} & \multirow{2}{*}{Method}  & \multicolumn{4}{c|}{\textbf{Early (25\%)}} & \multicolumn{4}{c}{\textbf{Middle (50\%)}} \\
              &  & R@1 & R@5 & R@10 &  MdR & R@1 & R@5 & R@10 &  MdR\\
    \midrule
    \multirow{4}{*}{\begin{tabular}{c}
    Feasible \\ Recipe\\ Retrieval
    \end{tabular}} & Random 
         & 1.8 & 8.6 & 15.8 & - 
         & 0.4 & 1.8 & 3.1 & -\\ 
    & UniVL~\cite{Luo2020UniVL}  
         & 3.4 &  10.3 & 17.2 &  56.0 
         & 3.4 &  10.3 & 17.2 &  56.0 \\
    & CLIP4Clip~\cite{Luo2021CLIP4Clip}
         & 3.4 & 6.8 & 13.7 & 60.0 
         & 3.4 & 3.4 & 10.3 & 94.0 \\ 
    & X-CLIP~\cite{Ma2022XCLIP} 
         & 3.4 & 10.3 & 13.7 & 111.0 
         & 0.0 & 3.4 & 3.4 & 569.0 \\
    \midrule
    \multirow{4}{*}{\begin{tabular}{c}
    Recipe \\ Stage \\ Identification
    \end{tabular}} & Random 
         & 6.3 & 31.6 & 63.3 & 8.0 
         & 6.3 & 31.6 & 63.3 & 8.0 \\
    & UniVL~\cite{Luo2020UniVL}  
         & 6.8 & 37.9 & 65.5 &  7.0 
         & 0.0 & 41.3 & 86.2 & 5.0 \\
    & CLIP4Clip~\cite{Luo2021CLIP4Clip}
         & 6.8 & 31.0 & 51.7 & 9.0 
         & 3.4 & 41.3 & 82.7 & 7.0 \\ 
    & X-CLIP~\cite{Ma2022XCLIP} 
         & 6.8 & 37.9 & 51.7 & 8.0 
         & 6.8 & 34.4 & 51.7 & 8.0 \\
    \bottomrule
    \end{tabular}
}
\end{table}

\begin{table*}
\caption{Online recipe retrieval (OnRR) performances of baseline models in late- (75\%) and full-stage (100\%) settings. The rows with `FT' of `\checkmark' show the results of models fine-tuned on the \comk dataset. Note that as the cooking stage progresses, random results in feasible recipe retrieval deteriorate due to the reduced number of feasible recipes.}
\label{tab:result_onrr_stage}
\centering
\scalebox{0.98}{
    \tabcolsep = 0.9mm
    \begin{tabular}{c|r|c|rrrr|rrrr}
    \toprule
    \multirow{2}{*}{Task} & \multicolumn{1}{c|}{\multirow{2}{*}{Method}} 
    & \multicolumn{1}{c|}{\multirow{2}{*}{FT}} 
    & \multicolumn{4}{c}{\textbf{Late (75\%)}} 
    & \multicolumn{4}{c}{\textbf{Full (100\%)}}\\
     &  &
    & R@1 & R@5 & R@10 &  MdR 
    & R@1 & R@5 & R@10 &  MdR\\
    \midrule
    \multirow{7}{*}{\begin{tabular}{c}
    Feasible \\ Recipe\\ Retrieval
    \end{tabular}} & Random & -
         & 0.0 & 0.0 & 0.0 & - 
         & 0.0 & 0.0 & 0.0 & - \\ 
    & UniVL~\cite{Luo2020UniVL} & 
         & 3.4 &  10.3 & 17.2 &  56.0 
         & 3.4 &  10.3 & 17.2 &  56.0 \\
    & UniVL~\cite{Luo2020UniVL} & \checkmark
         & 3.4 & 5.7 & 9.2 & 231.0 
         & 3.4 & 5.7 & 9.2 & 231.0 \\ 
    & CLIP4Clip~\cite{Luo2021CLIP4Clip} &
         & 3.4 & 3.4 & 10.3 & 85.0
         & 3.4 & 3.4 & 6.8 & 77.0 \\
    & CLIP4Clip~\cite{Luo2021CLIP4Clip} & \checkmark
         & 0.0 & 0.0 & 6.8 & 91.0 
         & 0.0 & 0.0 & 3.4 & 72.0 \\  
    & X-CLIP~\cite{Ma2022XCLIP} & 
         & 0.0 & 0.0 & 0.0 & 860.0
         & 0.0 & 0.0 & 0.0 & 911.0 \\ 
    & X-CLIP~\cite{Ma2022XCLIP} & \checkmark
         & 0.0 & 0.0 & 0.0 & 446.0
         & 0.0 & 0.0 & 0.0 & 366.0 \\
    \midrule
    \multirow{7}{*}{\begin{tabular}{c}
    Recipe \\ Stage\\ Identification
    \end{tabular}} & Random & -
         & 6.3 & 31.6 & 63.3 & 8.0
         & 6.3 & 31.6 & 63.3 & 8.0 \\
    & UniVL~\cite{Luo2020UniVL} & 
         & 0.0 & 41.3 & 86.2 & 5.0
         & 0.0 & 48.2 & 96.5 & 5.0 \\ 
    & UniVL~\cite{Luo2020UniVL}  & \checkmark
         & 6.8 & 44.8 & 86.2 & 5.0
         & 6.8 & 51.7 & 96.5 & 4.0 \\ 
    & CLIP4Clip~\cite{Luo2021CLIP4Clip} &
         & 3.4 & 20.6 & 79.3 & 7.0 
         & 0.0 & 24.1 & 93.1 & 7.0 \\ 
    & CLIP4Clip~\cite{Luo2021CLIP4Clip} & \checkmark
         & 6.8 & 48.2 & 93.1 & 5.0
         & 10.3 & 55.1 & 89.6 & 4.0 \\ 
    & X-CLIP~\cite{Ma2022XCLIP} & 
         & 10.3 & 41.3 & 93.1 & 6.0 
         &  6.8 & 58.6 & 89.6 & 4.0 \\ 
    & X-CLIP~\cite{Ma2022XCLIP} & \checkmark
         & 10.3 & 41.3 & 93.1 & 6.0 
         & 10.3 & 62.0 & 89.6 & 3.0 \\ 
    \bottomrule
    \end{tabular}
}

\end{table*}

\section{Additional Visual Examples on DVC-OV tasks}\label{app:sec:dvc_ov}

\begin{figure*}
    \centering
    \begin{subfigure}[b]{0.9\textwidth}
        \centering
        \includegraphics[width=\textwidth,trim={4cm 5cm 4cm 5cm},clip]{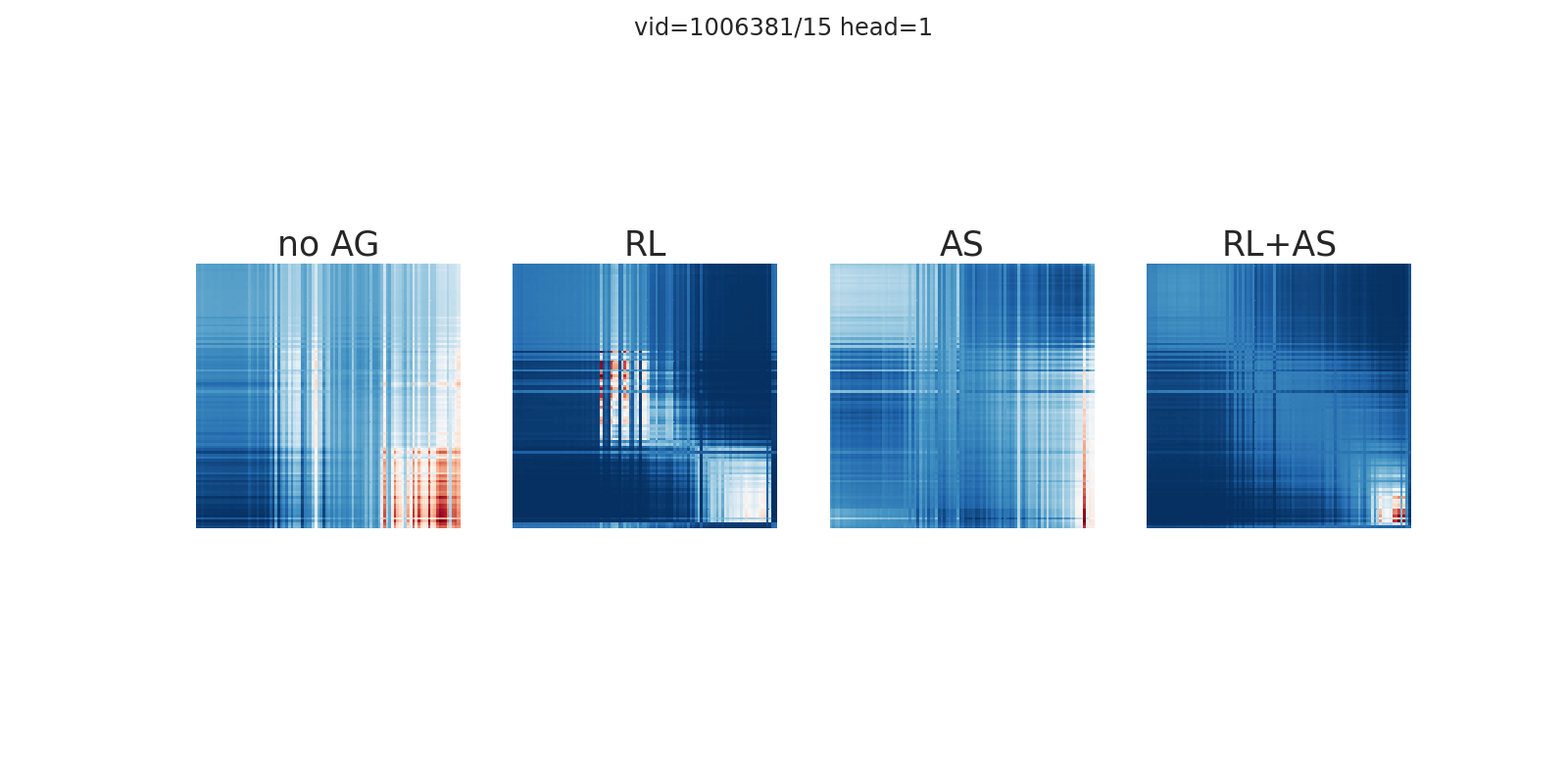}
    \end{subfigure}\\ \vspace{-1.0em}
    \begin{subfigure}[b]{0.9\textwidth}
        \centering
        \includegraphics[width=\textwidth,trim={4cm 5cm 4cm 6.67cm},clip]{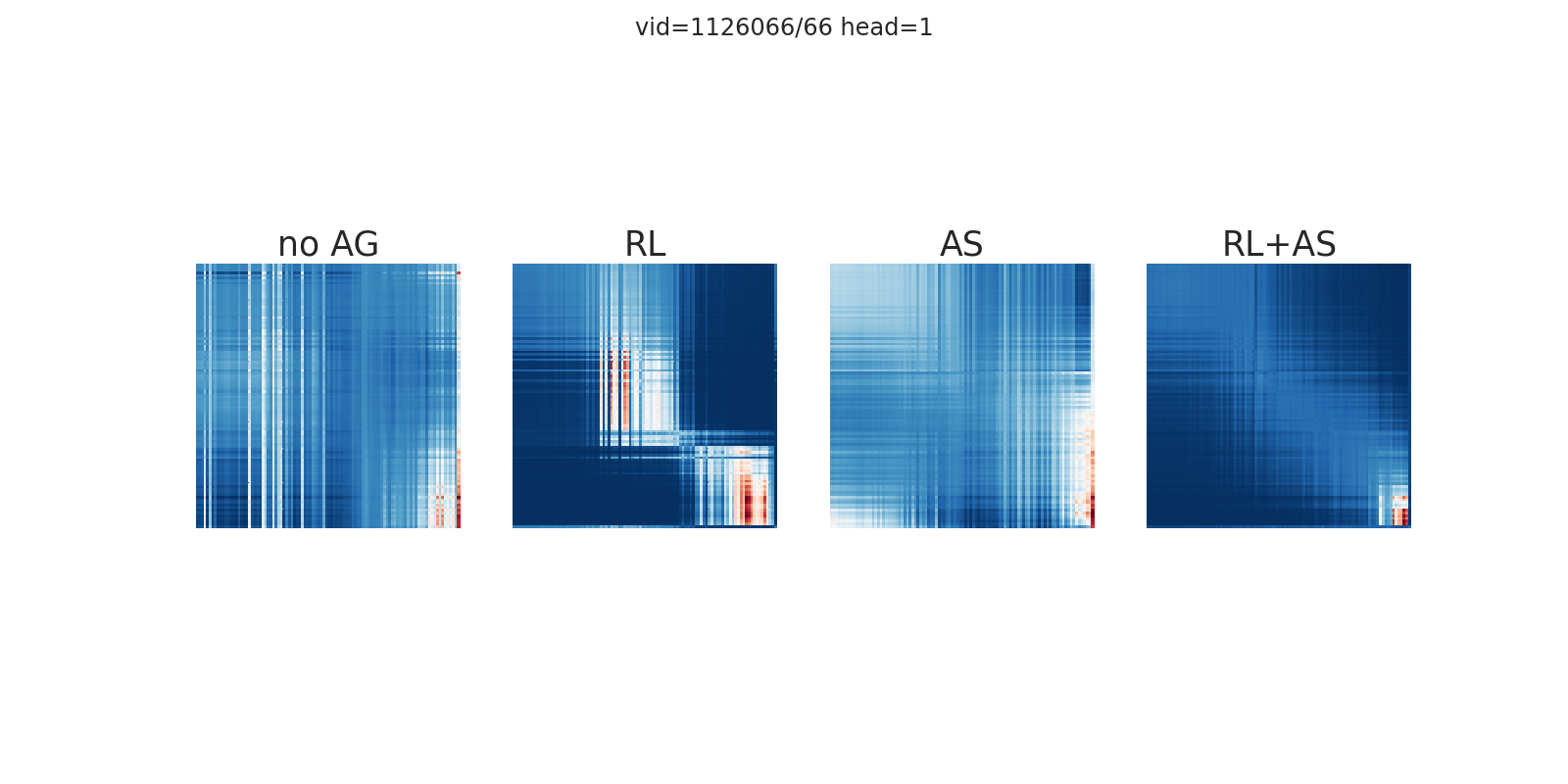}
    \end{subfigure}\\ \vspace{-1.0em}
    \begin{subfigure}[b]{0.9\textwidth}
        \centering
        \includegraphics[width=\textwidth,trim={4cm 5cm 4cm 6.67cm},clip]{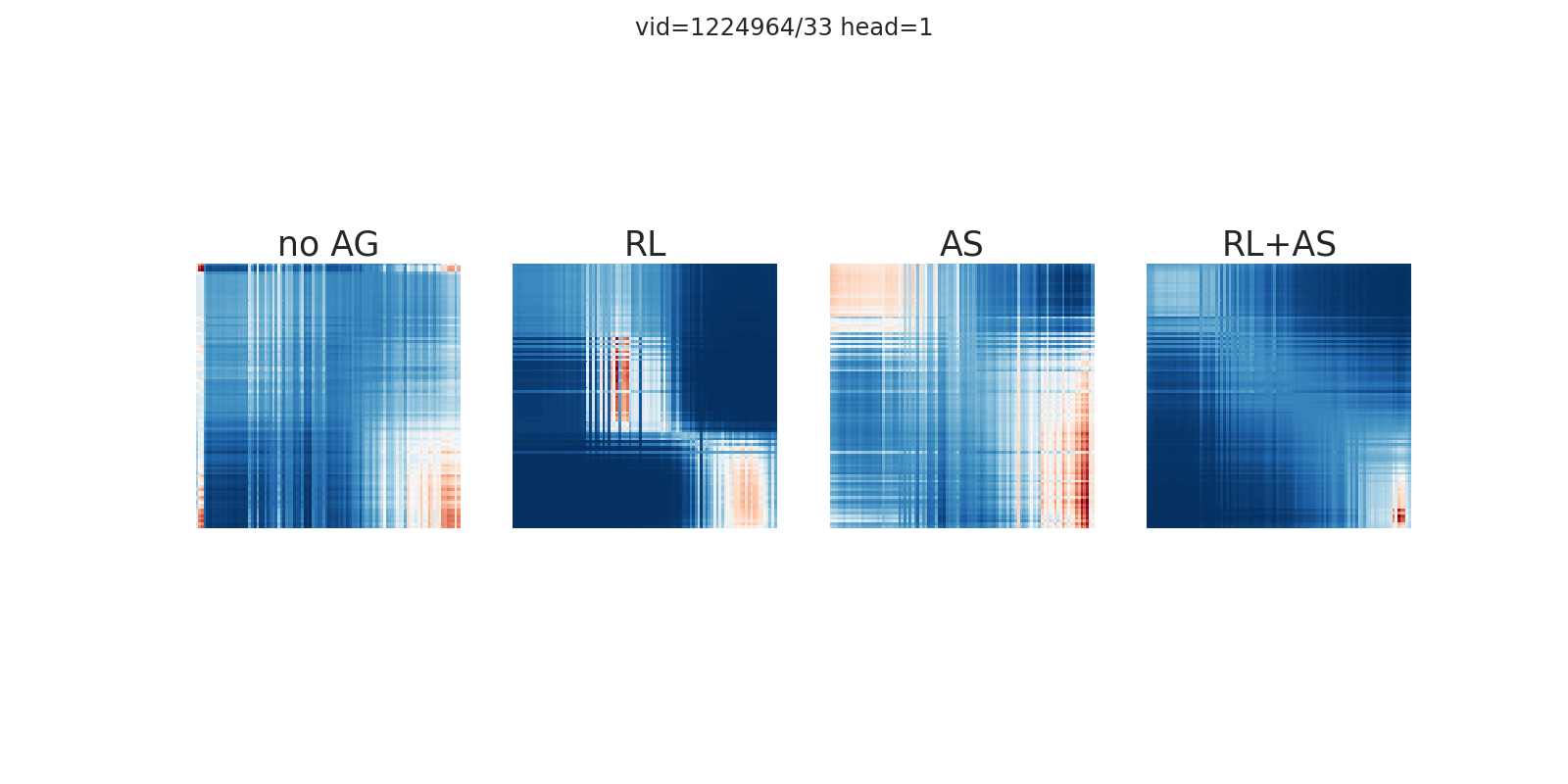}
    \end{subfigure}\\ \vspace{-1.0em}
    \begin{subfigure}[b]{0.9\textwidth}
        \centering
        \includegraphics[width=\textwidth,trim={4cm 5cm 4cm 6.67cm},clip]{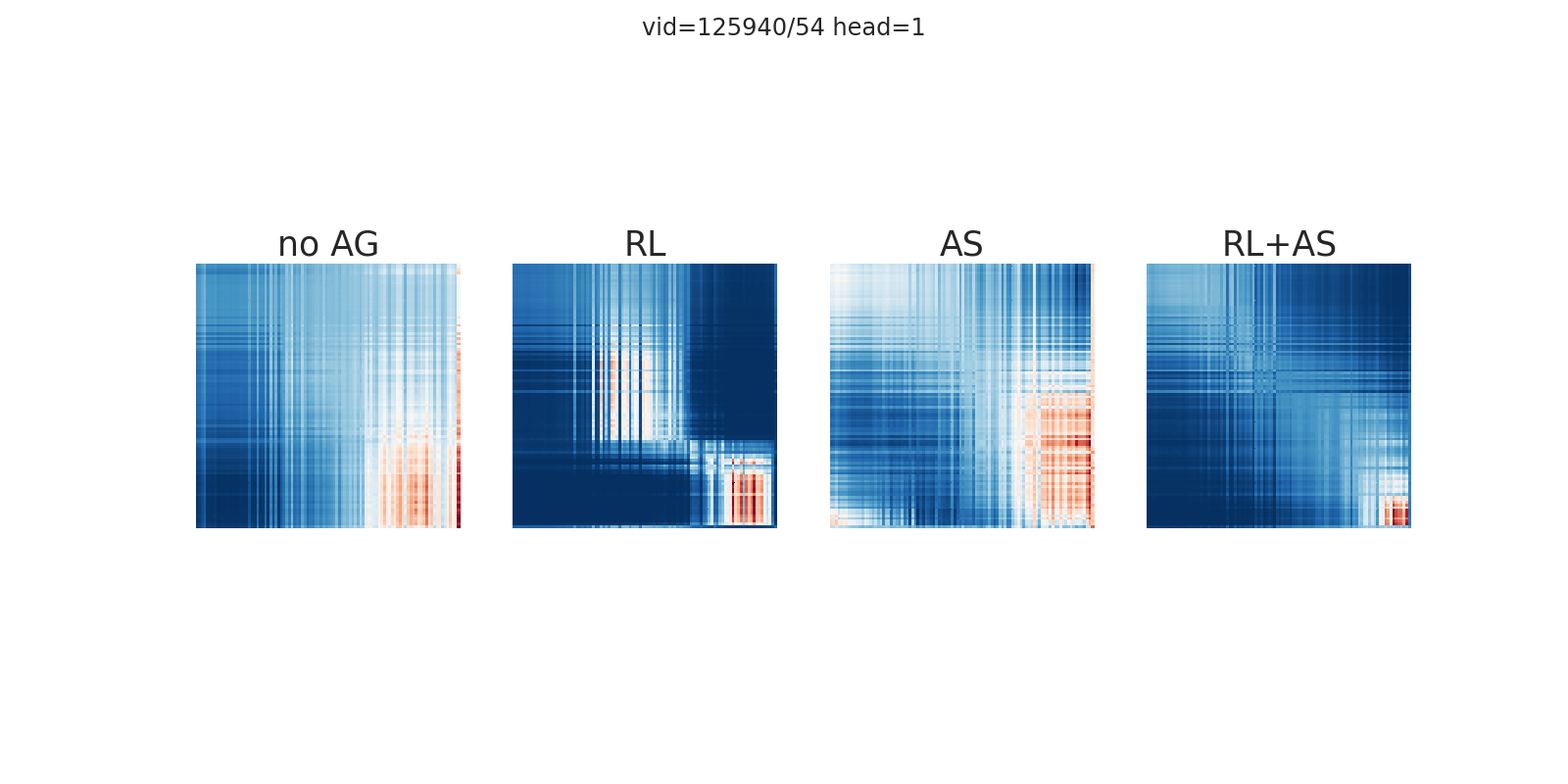}
    \end{subfigure}\\ \vspace{-1.0em}
    \begin{subfigure}[b]{0.9\textwidth}
        \centering
        \includegraphics[width=\textwidth,trim={4cm 5cm 4cm 6.67cm},clip]{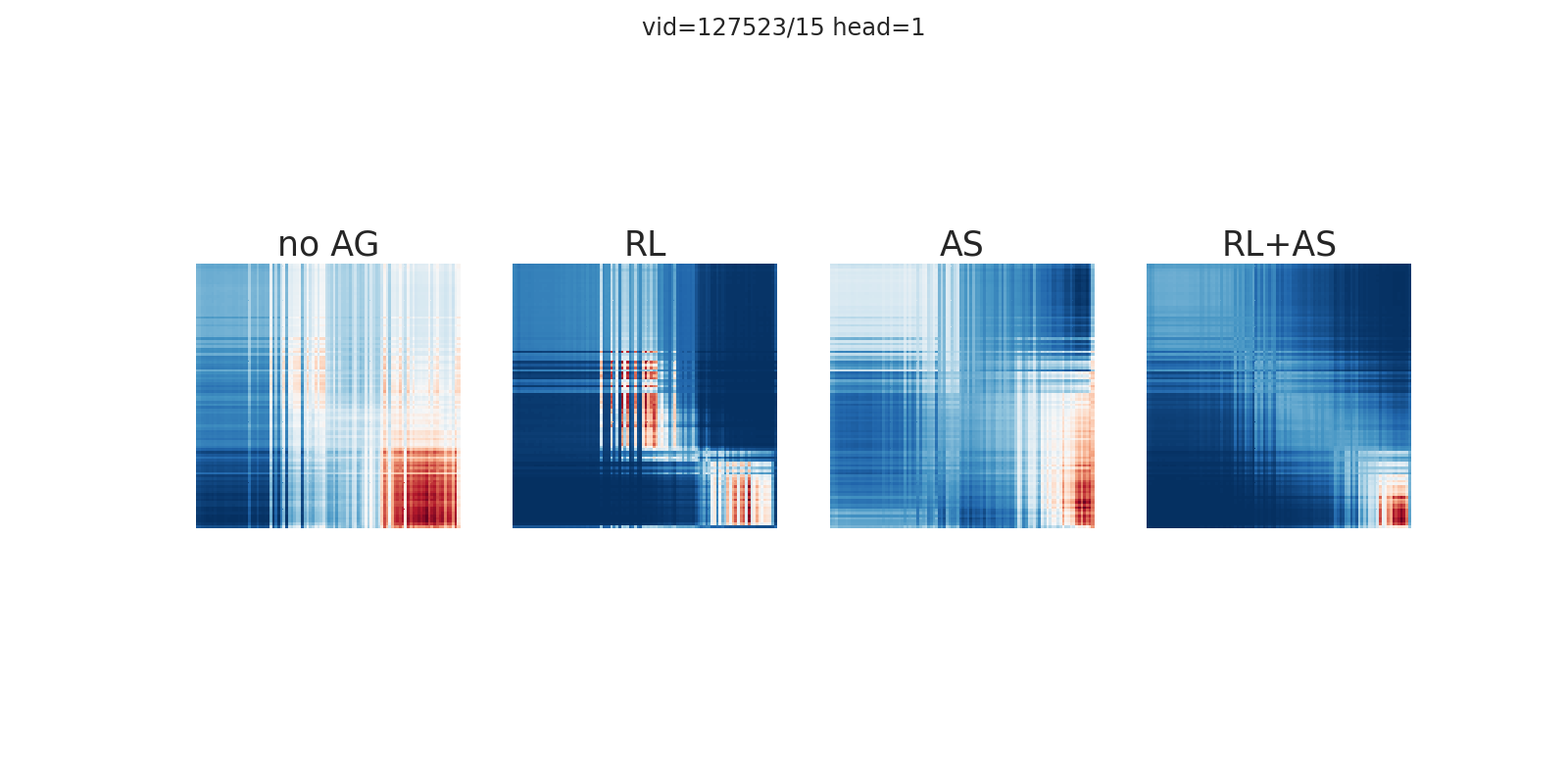}
    \end{subfigure}\\ \vspace{-1.0em}
    \begin{subfigure}[b]{0.9\textwidth}
        \centering
        \includegraphics[width=\textwidth,trim={4cm 5cm 4cm 6.67cm},clip]{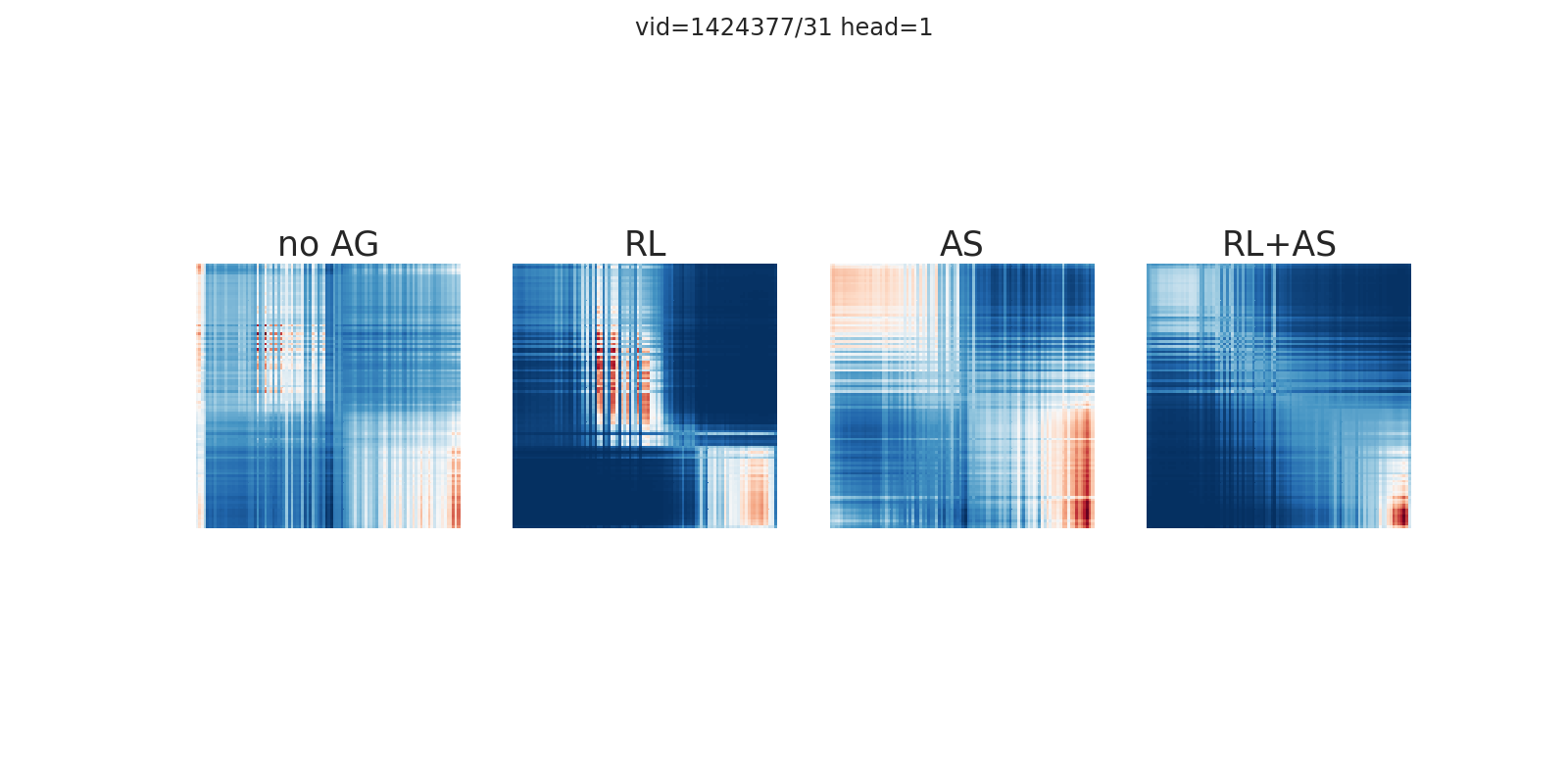}
    \end{subfigure}    
    \caption{
        Additional examples of attention of the first head at the last encoder layer.
        The red area indicates the high attention weights.
    }
    \label{fig:example-dvc-att}
\end{figure*}

The following examples are included to provide further insights and reinforce the points made in the main text.
Here, we present some more cases in \cref{fig:example-dvc-att}.
As with the other cases, we confirm that the combination of supervision connected related frames, using action graphs as relation labels (RL) and as attention supervision (AS).


\end{document}